\documentclass[acmlarge,screen]{acmart}
\usepackage{graphicx}
\usepackage{textcomp}
\usepackage{xcolor}
\usepackage{listings}
\usepackage{url}
\usepackage{xspace}
\usepackage{multirow}
\usepackage{balance}
\usepackage{ctable}
\usepackage{listings}
\usepackage{color}
\usepackage{graphicx}
\usepackage[nobiblatex]{xurl}

\usepackage{multirow}

\usepackage{caption}
\usepackage{subcaption}

\usepackage{cleveref}
\usepackage{acronym}
\usepackage{fancybox}
\usepackage{verbatim}
\usepackage[normalem]{ulem}
\usepackage{float}
\usepackage{newfloat}
\usepackage{mdframed}
\usepackage{enumitem}
\usepackage{booktabs}
\usepackage{pifont}

\usepackage{amsmath}
\usepackage{amsthm}
\usepackage{adjustbox}
\usepackage{hyperref}
\usepackage{academicons}

\usepackage{algorithm}
\usepackage{algorithmic}

\usepackage{tcolorbox}

\usepackage{colortbl}  
\usepackage{xcolor}
\usepackage{array}   

\definecolor{red}{RGB}{255,0,0}
\definecolor{green}{RGB}{18,220,168}

\newcommand{\ie}{i.e.,\xspace}
\newcommand{\eg}{e.g.,\xspace}

\newcommand{\myorcid}[1]{\href{https://orcid.org/#1}{\textcolor[HTML]{A6CE39}{\aiOrcid}}}

\definecolor{pkured}{RGB}{192,0,0}

\newcommand{\changes}[1]{\textcolor{black}{#1}}
\newcommand{\revised}[1]{\textcolor{black}{#1}}

\newcommand{\tool}{\textsc{Moss}\xspace}

\newcommand{\eat}[1]{}
\acrodef{wp}[WP]{Website Fingerprinting}
\acrodef{apt}[APT]{Advanced Persistent Threats}
\acrodef{lol}[LotL]{Living-Off-The-Land}
\acrodef{ids}[IDS]{Intrusion Detection System}
\acrodef{vae}[VAE]{Variational AutoEncoder}
\acrodef{re}[RE]{reconstruction error}
\acrodef{sv}[SV]{Stableness Value}
\acrodef{as}[AS]{anomaly score}
\acrodef{gas}[GAS]{graph anomaly score}
\acrodef{etw}[ETW]{Event Tracing for Windows}
\acrodef{e3}[E3]{Engagement 3}
\acrodef{e5}[E5]{Engagement 5}
\acrodef{ttp}[TTPs]{Tactics, Techniques, and Procedures}
\acrodef{hsg}[HSG]{High-level Scenario Graph}
\acrodef{nlp}[NLP]{Natural Language Processing}
\acrodef{dg}[DG]{Detection Graph}
\acrodef{poi}[POI]{Point of Interest}
\acrodef{iv}[IV]{Important Value}
\acrodef{sg}[SG]{Suspicious Graph}
\acrodef{mttd}[MTTD]{Mean Time to Detect}
\acrodef{soc}[SOC]{Security Operations Center}

\newlength{\MaxSizeOfLineNumbers}%
\definecolor{keywordcolor}{rgb}{0.8,0.1,0.5}
\definecolor{lightlightgray}{gray}{.96}
\definecolor{lightgray}{gray}{.925}
\definecolor{medlightgray}{gray}{0.7}
\definecolor{medgray}{gray}{0.4}
\definecolor{darkgray}{gray}{0.35}
\definecolor{nearblack}{gray}{0.15}

\crefname{component}{Component}{Components}

\newtheorem{defn}{Definition}

\AtBeginDocument{%
  }

\setcopyright{acmlicensed}
\acmJournal{IMWUT}
\acmYear{2025} \acmVolume{9} \acmNumber{1} \acmArticle{211} \acmMonth{3}\acmDOI{10.1145/3712274}

\begin{document}

\title{\tool: Proxy Model-based Full-Weight Aggregation in Federated Learning with Heterogeneous Models}


\author{Yifeng Cai}
\email{caiyifeng@pku.edu.cn}
\orcid{0000-0002-0049-6670}
\affiliation{%
  \institution{MOE Key Lab of HCST (PKU), School of Computer Science, Peking University}
  \city{Beijing}
  \country{China}
  \postcode{100871}
}

\author{Ziqi Zhang}
\email{ziqi24@illinois.edu}
\orcid{0000-0001-8493-0261}
\affiliation{%
  \institution{Department of Computer Science, University of Illinois Urbana-Champaign}
  \city{Urbana}
  \state{IL}
  \country{USA}
  \postcode{61801}
}

\author{Ding Li}
\email{ding_li@pku.edu.cn}
\orcid{0000-0001-7558-9137}
\authornote{Ding Li and Yao Guo are the corresponding authors.}
\affiliation{%
  \institution{MOE Key Lab of HCST (PKU), School of Computer Science, Peking University}
  \city{Beijing}
  \country{China}
  \postcode{100871}
}

\author{Yao Guo}
\authornotemark[1]
\email{yaoguo@pku.edu.cn}
\orcid{0000-0001-5064-5286}
\affiliation{%
  \institution{MOE Key Lab of HCST (PKU), School of Computer Science, Peking University}
  \city{Beijing}
  \country{China}
  \postcode{100871}
}

\author{Xiangqun Chen}
\email{cherry@pku.edu.cn}
\orcid{0000-0002-7366-5906}
\affiliation{%
  \institution{MOE Key Lab of HCST (PKU), School of Computer Science, Peking University}
  \city{Beijing}
  \country{China}
  \postcode{100871}
}

\renewcommand{\shortauthors}{Cai et al.}

\begin{abstract}
Modern Federated Learning (FL) has become increasingly essential for handling highly heterogeneous mobile devices. Current approaches adopt a partial model aggregation paradigm that leads to sub-optimal model accuracy and higher training overhead. In this paper, we challenge the prevailing notion of partial-model aggregation and propose a novel "full-weight aggregation" method named \tool, which aggregates all weights within heterogeneous models to preserve comprehensive knowledge.  Evaluation across various applications demonstrates that \tool significantly accelerates training, reduces on-device training time and energy consumption, enhances accuracy, and minimizes network bandwidth utilization when compared to state-of-the-art baselines.
\end{abstract}



\maketitle

\section{Introduction}
\label{sec:intro}

Federated learning (FL)~\cite{mcmahan2017communication, zhao2018federated} has been applied to various mobile/IoT applications, such as human activity recognition~\cite{ouyang2021clusterfl,tu2021feddl}, mobile vision~\cite{li2021hermes,li2021fedmask}, and voice assistance~\cite{leroy2019federated}. The key challenge for adopting FL in modern mobile/IoT environments is how to deploy models on heterogeneous devices while achieving desirable accuracy. Conventional FL requires that all mobile devices have the same model~\cite{mcmahan2017communication, li2020federated, li2020federated2}. However, this requirement is not practical for mobile/IoT environments since devices can be highly heterogeneous~\cite{horvath2021fjord}.

The heterogeneity of devices leads to heterogeneity in the deployment of deep neural network (DNN) models. First, manufacturers and developers can select or customize design models to fit hardware characteristics~\cite{lee2020hardware,hussain2022design,hao2019fpga}, resulting in heterogeneous model deployments across devices. Second, developers may choose models with varying architectures and parameter sizes for devices with different computational capacities to optimize performance~\cite{caldas2018expanding}. Moreover, in the context of model heterogeneity, employing FL systems becomes imperative. This enables collaborative model training while ensuring privacy protection, ultimately enhancing model accuracy.
Therefore, to address the heterogeneity between devices, researchers have developed multiple solutions that support developers to deploy heterogeneous models with different architectures to different devices in the FL system~\cite{li2021lotteryfl,li2021hermes,li2021fedmask,deng2022tailorfl,luo2021fedskel,diao2020heterofl,horvath2021fjord,li2019fedmd,lin2020ensemble,itahara2020distillation,cheng2021fedgems,cho2022heterogeneous}.

\begin{figure*}[!tb]
    \centering
    \includegraphics[width=0.85\linewidth]{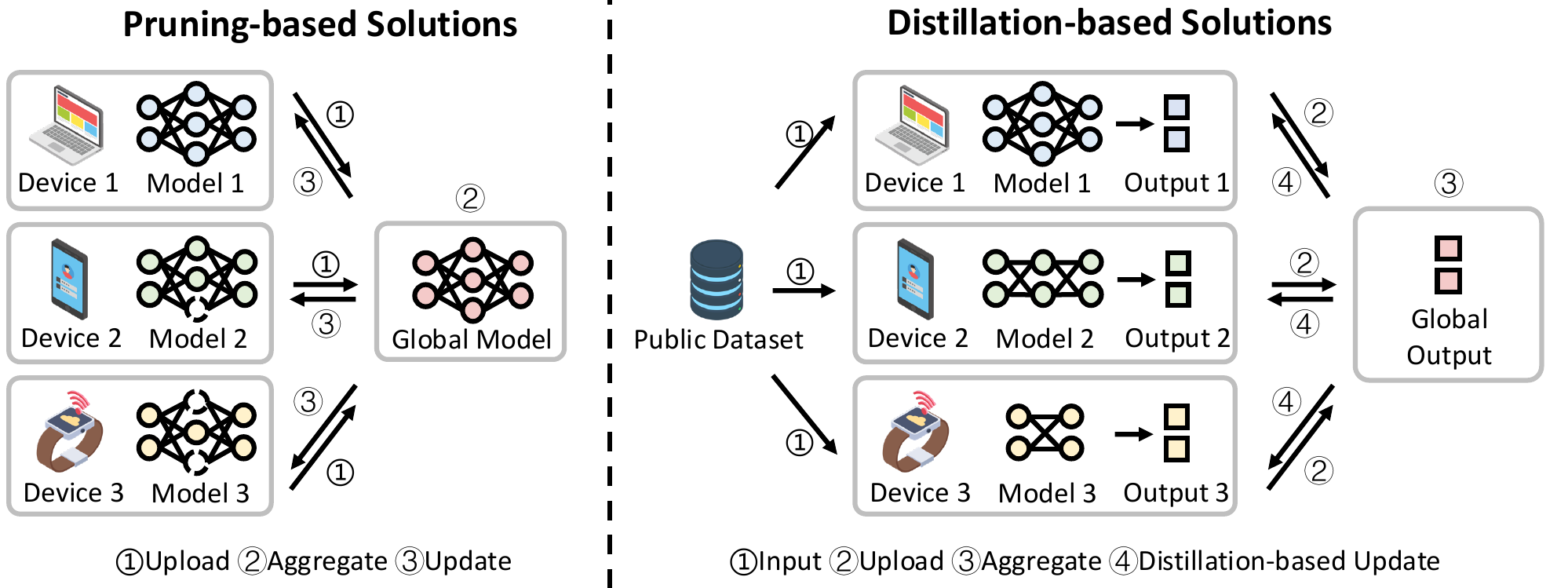}
	\caption{Illustration of pruning-based solutions and distillation-based solutions.}
	\label{fig:existing}
\end{figure*}

The key challenge for heterogeneous FL is \textit{how to aggregate weights across models with different architectures}. Unlike conventional homogeneous FL, it is impossible to directly have the conventional full-weight aggregation (i.e., model aggregation based on the index of weight) between heterogeneous models. 
To address this challenge, researchers developed several approaches following the three types of solutions~\cite{fan2023model}: pruning-based solutions~\cite{li2021lotteryfl,li2021hermes,li2021fedmask,deng2022tailorfl,luo2021fedskel,diao2020heterofl,horvath2021fjord}, distillation-based solutions~\cite{li2019fedmd,lin2020ensemble,itahara2020distillation,cheng2021fedgems,cho2022heterogeneous}, and alternative solutions~\cite{litany2022federated,rapp2022distreal,yao2021fedhm}. We illustrate the common design of pruning-based and distillation-based solutions in Figure~\ref{fig:existing}. 

Specifically, pruning-based solutions~\cite{li2021lotteryfl,li2021hermes,li2021fedmask,deng2022tailorfl,luo2021fedskel,diao2020heterofl,horvath2021fjord} capitalize on the concept of model pruning, which posits that existing models are over-parameterized. Therefore, these solutions first initialize an original model and then remove some of the less crucial weights within the model to tailor it to adapt the computational capabilities of the devices. During the aggregation phase of federated learning, only the weights of the pruned sub-models in heterogeneous models are aggregated and updated; the weights that have been pruned are not updated. However, when models are highly heterogeneous, these solutions may fail. If the original model is excessively pruned to adapt devices with lower computational power~\cite{10023963}, it can disrupt the original model architecture, leading to significant reductions in accuracy and preventing the model from converging~\cite{liu2018rethinking,zhu2017prune}. Moreover, many devices may have model architectures that are custom-designed for specific devices~\cite{lee2020hardware,hussain2022design,hao2019fpga}, making it difficult to find a consistent original architecture across different device models, thus pruning-based solutions have significant limitations.

Distillation-based solutions~\cite{li2019fedmd,lin2020ensemble,itahara2020distillation,cheng2021fedgems,cho2022heterogeneous} are based on the principle that the output dimensions of heterogeneous models within the same task are identical. These solutions first utilize a public dataset as input to generate outputs from heterogeneous models, which are then aggregated. Subsequently, model distillation techniques are employed to compute the differences between the aggregated outputs and the original outputs, thereby optimizing each heterogeneous model. However, these solutions overlook the knowledge in the intermediate layers of the models, and the outputs they utilize contain less information~\cite{romero2014fitnets,heo2019comprehensive,tung2019similarity}, which extends the convergence time of training. For devices with limited computational capabilities, this significantly increases computational overhead and reduces the model accuracy.

Alternative solutions~\cite{fan2023model} also tried to achieve FL with heterogeneous models. HAFL-GHN~\cite{litany2022federated} employs a Graph HyperNetwork to adapt to heterogeneous client architectures while preserving privacy. However, it does not establish correspondences between different architectures and avoids aggregating layers without matches, making it hard to apply in the real-world. DISTREAL~\cite{rapp2022distreal} introduces a distributed resource-aware learning pattern that allows clients to adapt by omitting dynamic filters. Yet, like pruning-based methods, it requires an initial large model, limiting its use in highly heterogeneous environments. FEDHM~\cite{yao2021fedhm} decomposes model parameters into shared and individual components, reducing communication costs. However, it incurs significant computational overhead and demands shared parameters that exceed the capabilities of low-end devices, hindering its adaptability in diverse settings.

In summary, all of the solutions suffer from sub-optimal accuracy and training efficiency. The reason is that these approaches take a partial model aggregation paradigm, thus can only aggregate a subset of the heterogeneous models. This paradigm omits the knowledge learned from other unaggregated parts of different heterogeneous models, which not only downgrades the accuracy of the trained model but also prolongs the training process since extra effort is required to recover the lost knowledge, leading to more energy and time consumption.

To address this problem, we propose a novel \textit{full-weight aggregation} paradigm for heterogeneous FL, which allows incorporating the full knowledge of heterogeneous models. Our insight is that full-weight aggregation (e.g. FedAvg~\cite{mcmahan2017communication}) has shown its effectiveness in maximizing model accuracy and reducing training overhead in homogeneous FL. Thus, if we can find an effective way to map the weight correspondence between heterogeneous models, we can simulate full weight aggregation in heterogeneous FL to improve model accuracy and reduce training overhead. 

The key challenge for our \textit{full-weight aggregation} is how to find the correct weight mapping between models with different architectures. To this end, we propose a proxy model-based approach that first converts heterogeneous models to homogeneous proxy models and then performs full-weight aggregation on the homogeneous proxy models. In this way, our approach can fully aggregate the knowledge learned from all mobile/IoT devices, allowing higher model accuracy and lower training overhead. 

We implement our approach as a tool named \tool and evaluate the performance of \tool with hundreds of devices in three typical applications of heterogeneous FL, including image classification, speech recognition, and human activity recognition, to verify the effectiveness of \tool on real-world data.
Experimental results show that \tool can speed up the FL training process by up to 63.6 percentage points, compared to baselines. For the efficiency of mobile device computation, \tool can decrease the training time on the device by $62.9\%$ and decrease the energy consumption by 6.1$\times$. For the accuracy of the device models, \tool can improve the accuracy by 8.6 percentage points. More importantly, our evaluation shows that \tool consumes less total network bandwidth than 9 out of 12 existing state-of-the-art baselines. These data strengthen our argument for using a full-weight aggregation method.

In summary, the main contributions of this paper are as follows. 

\begin{itemize}
     \item We propose the \textit{full-weight} aggregation paradigm for FL with heterogeneous models that reduces the overall training overhead and improves the model accuracy.
    \item We design and implement \tool, the first FL scheme for heterogeneous models based on full-weight aggregation.
    \item We conduct extensive experiments on multiple real-world applications with different heterogeneous model architectures. The results demonstrate the performance of \tool with respect to the model accuracy and overall system overhead.
\end{itemize}

\section{Background and Motivation}
\label{sec:motivation}

\subsection{FL with Heterogeneous Models on Mobile/IoT Devices}
\changes{
FL is a common technique for collaborative model training across devices that enhances prediction accuracy while preserving user data privacy. FL typically involves three steps: 1) each device trains a model, and uploads the model to the server; 2) the server conducts full-weight model aggregation based on the index of device models' weights; 3) the server deploys the aggregated model to devices to achieve model update. This paradigm of FL can be applied in various scenarios, including mobile/IoT devices.}

Implementing FL on mobile/IoT devices requires adapting to significant differences in computational capabilities. Due to computational constraints, not all mobile/IoT devices can deploy the same model architecture. However, these devices must still participate in the FL system to improve model accuracy and thus provide optimal service to users.

Therefore, the models deployed in mobile/IoT devices are heterogeneous. First, the model architecture may be customized by device manufacturers based on their hardware characteristics~\cite{lee2020hardware,hussain2022design,hao2019fpga}. For example, existing research involves manually or using techniques such as Neural Architecture Search (NAS)~\cite{zoph2016neural,tan2019mnasnet} to design device-specific DNN models, optimizing training and inference efficiency on mobile/IoT devices. Additionally, developers may select model architectures based on the computational capabilities of the devices~\cite{liu2022no}. For instance, in IoT devices like smart TVs, due to limitations in RAM and CPU~\cite{10023963}, only models with 10K parameters might be deployable; whereas, the latest smartphones can easily train models with over 100 million parameters.

Simply using a traditional FL framework to deploy homogeneous models to such heterogeneous devices does not address these issues. Employing lightweight model architectures significantly reduces model accuracy, diminishing the performance that advanced smartphones are capable of, thus degrading user experience. Conversely, deploying complex and modern model architectures would make it impossible for low-end devices to sustain the models, thus excluding them from modern AI services~\cite{liu2022no}. Therefore, FL on mobile/IoT devices should deploy heterogeneous models adapted to the specific characteristics of each device, maximizing computational resources to enhance model accuracy as much as possible.

Overall, designing an effective and efficient framework for FL with heterogeneous models on mobile/IoT devices should achieve the following goals:

\textbf{Goal 1.} Improve accuracy and reduce training overhead. The basic objective of FL on mobile/IoT devices is to enhance accuracy and accelerate model convergence. Therefore, the primary goal is to speed up convergence, thereby reducing on-device overhead.

\textbf{Goal 2.} No additional computation overhead. The computational power and energy of devices are already limited, and it should not be necessary to incur additional overhead outside of model training to complete the FL process.

\textbf{Goal 3.} Compatible for arbitrary model architectures. In mobile/IoT devices, model architectures are likely highly heterogeneous. A robust FL framework with heterogeneous models should be as compatible as possible with arbitrary model architecture.

\subsection{Limitation of Existing Solutions}

When models are heterogeneous, conventional full-weight model aggregation cannot be directly performed due to the lack of a corresponding relationship between weights based on the index. Thus, researchers have proposed three main solutions for FL with heterogeneous models: pruning-based solutions, distillation-based solutions, and alternative solutions.

\subsubsection{Limitation on Pruning-based Solutions}
Pruning-based solutions~\cite{li2021lotteryfl,li2021hermes,li2021fedmask,deng2022tailorfl,luo2021fedskel,diao2020heterofl,horvath2021fjord} leverage model pruning techniques, operating on the premise that models are often over-parameterized. Thus, non-essential weights are trimmed to scale down the model to match the computational capabilities of the device. As illustrated in the left part of Figure~\ref{fig:existing}, during local model training, pruning-based solutions select a subset of the original model for training based on device capability and weight importance. During aggregation, the sub-model can be mapped back to the original model through the index recorded during pruning to complete the model aggregation.

However, pruning-based solutions face several limitations. First, these solutions require the use of the same original model for pruning. As discussed in the previous section, manufacturers are likely to design models customized to hardware specifications~\cite{lee2020hardware,hussain2022design,hao2019fpga}. Therefore, across different devices, it is challenging to match the index between heterogeneous models, which complicates the model aggregation process. Second, excessive pruning can severely degrade performance if too many parameters are lost or critical architectures are compromised~\cite{zhu2017prune,liu2018rethinking}. For example, when a ResNet model with 11.7M parameters~\cite{he2016deep} needs to be deployed on a Smart TV, at least 95\% of the weights must be removed to compress the parameters to fewer than 10K. This level of reduction can disrupt the model architecture, leaving three convolutional layers with as few as four channels and disabling some residual architectures, thereby preventing the model from converging and significantly affecting the training efficiency and overall accuracy of the FL system. Therefore, this method is not suitable for real-world applications on mobile/IoT devices.

\subsubsection{Limitation of Distillation-based Solutions}
As shown in the right part of Figure~\ref{fig:existing}, distillation-based solutions~\cite{li2019fedmd,lin2020ensemble,itahara2020distillation,cheng2021fedgems,cho2022heterogeneous} draw inspiration from model distillation techniques, processing inputs from a public dataset across all device models to obtain outputs of the same dimensions, which are then aggregated. Then, they can calculate the difference between aggregated outputs and original outputs as the loss to update the device models. The strength of these solutions lies in their compatibility with models of any architecture.

However, aggregating outputs leads to a loss of knowledge in intermediate layers, diminishing accuracy and prolonging the FL training rounds. Knowledge exists in every layer of a DNN model, with intermediate layers containing rich information~\cite{romero2014fitnets,heo2019comprehensive,tung2019similarity}. Therefore, retaining knowledge in intermediate layers is crucial during aggregation. Nonetheless, due to the operation of global pooling layers—which average features from the last convolutional layer to reduce dimensions—the output aggregation process predominantly captures limited class knowledge ~\cite{9340578,cho2019efficacy} and results in the loss of detailed information from intermediate layers. This inevitable knowledge loss further compromises model accuracy and extends the number of training rounds required to achieve convergence.

\subsubsection{Limitation of Alternative Solutions}
\revised{
Despite the pruning-based techniques and distillation-based techniques, several solutions have emerged to address FL with heterogeneous models. HAFL-GHN~\cite{litany2022federated} employs a Graph HyperNetwork to handle heterogeneous device architectures, enabling privacy while adapting to diverse architectures. However, it lacks a clear method for establishing correspondences between different model architectures. Moreover, for layers without corresponding matches, it chooses not to aggregate them. This limitation makes it difficult to align layers between DNNs with distinct parameters, layers, and architectures, hindering its applicability in real-world deployments where various architectures coexist.}

\revised{
DISTREAL~\cite{rapp2022distreal} utilizes a distributed resource-aware learning pattern, allowing devices to adapt independently by dropping dynamic filters to reduce training complexity. However, it operates similarly to pruning-based methods, necessitating the drop from an initial large model. This requirement restricts its effectiveness in scenarios where devices are highly heterogeneous, as not all devices can effectively support such large initial models.}

\revised{
FEDHM~\cite{yao2021fedhm} decomposes model parameters into shared and individual components, effectively reducing communication costs. However, it incurs significant computational overhead due to matrix composition, which can be a barrier for many systems. Additionally, the requirement for shared model parameters to exceed the computational capabilities of low-end devices limits its deployment in environments where resources are constrained, further challenging its adaptability to heterogeneous settings.}

\section{Design of \tool}
\label{sec:approach}

In this paper, we propose \tool, the first approach that achieves full-weight aggregation for heterogeneous FL.
There are three major design goals for \tool: 1, improve the accuracy and reduce the number of training rounds, thereby reducing the training overhead on the devices. 2, do not introduce any additional computation overhead on devices. 3, compatible for arbitrary model architectures, in order to increase the generalizability of \tool. Besides these three primary goals, we also aim to achieve the general goals of FL, namely keeping the users' privacy and avoiding extensive data uploading. To achieve the general goals of FL, we adopt the standard FL solutions that only upload model weights to the server for model aggregation~\cite{mcmahan2017communication,gao2022survey}. Overall, our approach keeps the users' privacy and the amount of uploaded data at the same level as existing approaches~\cite{mcmahan2017communication,liu2021distfl,liu2021pfa,zhang2023fedslice,fedbalancer,lai2021oort}.

We follow existing approaches to assume that the server can leverage public datasets to facilitate FL aggregation~\cite{lin2020ensemble,cho2022heterogeneous, zhao2018federated,zhang2023fedslice,cao2021fltrust,chang2019cronus,10.1145/3503161.3548764,li2020learning}. This assumption is particularly true today as there are many high-quality public datasets in the market~\cite{deng2009imagenet}. We assume that the server has no access to the local data of devices, thereby ensuring the privacy of users.

\subsection{Approach Overview}

\begin{figure}[htb]
    \centering
    \includegraphics[width=1\linewidth]{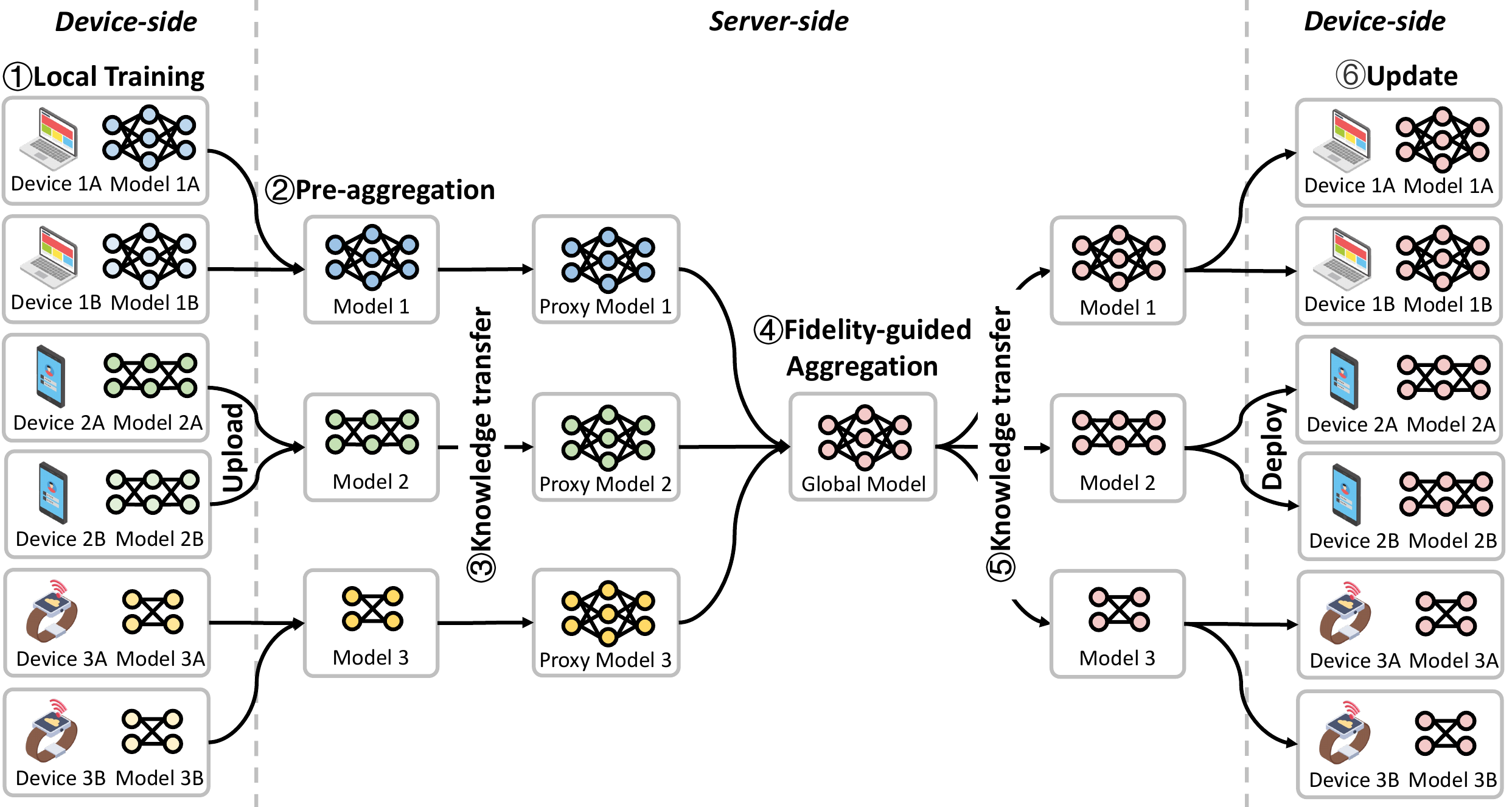}
	\caption{Framework of \tool.}
	\label{fig:framework}
\end{figure}

We first introduce the overview of \tool, which is composed of three components:
proxy model construction (PROM), weight-wise knowledge transfer (WIRE), and
fidelity-guided aggregation (FILE). The first component
is conducted only once during the system setup phase. The second and third
components are executed iteratively during the training process, which is shown
in Figure~\ref{fig:framework}.

The first component of \tool is proxy model construction (PROM). The goal of this
component is to alleviate the amount of computation cost on the server side and
further reduce the waiting time of devices. Specifically, we construct a series
of proxy models with homogeneous architectures for each type of device model.
The proxy models serve as the intermediate models for fine-grained knowledge
transfer and aggregation. By constructing PROM, \tool only needs to identify the mapping correspondences between each distinct architecture and its corresponding proxy model. Subsequently, aggregating all homogeneous proxy models effectively accomplishes full-weight aggregation.

The second component of \tool is weight-wise knowledge transfer (WIRE). The goal of
this component is to establish the weight-wise mapping relationship between the proxy model and the device model. Specifically, we leverage meta-learning to effectively learn the knowledge transfer mapping at both the layer-to-layer and neuron-to-neuron levels.

The third component is fidelity-guided aggregation (FILE). The goal of this component
is to optimize the aggregation process and facilitate convergence. Specifically,
we measure the distance between the proxy model and the original as the fidelity
of the proxy models. The fidelity serves as the importance score to aggregate
the proxy models. In this way, the proxy models that align better with the
original models have higher weights during aggregation, thus the knowledge in
the origin model is more accurately transferred to the aggregated models.

The training pipeline of \tool is displayed in Figure~\ref{fig:framework}, which
consists of six steps. At first, each device trains the local model with its
private data (\ding{172}). Then, devices upload heterogeneous models and the
server conducts pre-aggregation with the same architectures (\ding{173}). After
that, the server transfers knowledge from the pre-aggregated models to the proxy
models via the fine-grained mapping (\ding{174}). Next, the server aggregates
the proxy models by their fidelities (\ding{175}). Then, the server transfers
knowledge from the aggregated global model to each distinct model architecture
(\ding{176}). Finally, the updated models are sent back to the devices
(\ding{177}). The workflow iterates for multiple times until a desired
performance or a predefined number of iterations. After training, each device receives a well-trained model that fits to its computation ability and is enriched with comprehensive federated knowledge.

\subsection{Problem Definition}

The concept of full-weight aggregation with heterogeneous models can be interpreted as follows: 

\begin{defn}
\label{defn:problem}
\textbf{(Full-weight aggregation on heterogeneous models.)}
Given a set of heterogeneous models, full-weight aggregation should aggregates all weights of the models to compute the weights of the global model. No weights are discarded during the aggregation process.
\end{defn}

In our framework, there are two parties: multiple devices and one server. Each
device has a private dataset to train the local model. and the server is
responsible for aggregating the models from devices. We focus on the
heterogeneous scenario, where the devices are equipped with diverse model
architectures. Specifically, we assume that there are $N$ types of model
architectures, and each type of model architecture has $K_N$ devices. For each
type of model architecture, the devices are trained with the same model
architecture but different private datasets. Formally, we use $M^i_j$ to denote
the model on $j_{th}$ device of the $i_{th}$ model architecture, where
$i\in\{1,\dots,N\}, j\in\{1,\dots,K_N\}$. We also use $D^i_j$ to denote the
private dataset used to train $M^i_j$. 
\revised{
Algorithm~\ref{alg:moss} provides the concrete process
of \tool. In the subsequent sections, we will provide a
detailed description of each key design.}

\begin{algorithm}[h]
\color{black}
\caption{\revised{Moss}}
\label{alg:moss}
\begin{flushleft}
\textbf{Server-side:}
\end{flushleft}
\begin{algorithmic}[1]
    \STATE Initialize global proxy models $PM^1, PM^2, \dots, PM^N$ for each device model architecture \texttt{//PROM}
    \FOR{each round $t = 1, 2, \dots, T$}
        \FOR{each device type $i \in \{1,\dots,N\}$ in parallel}
            \STATE Broadcast corresponding latest device model $M^i$ to the $i$-th type devices
            \FOR{each device $j \in \{1,\dots,K_i\}$ in parallel}
                \STATE Receive local model updates $M^i_j$ from device $j$
            \ENDFOR
            \STATE $M^i = FedAvg(M^i_1,\dots,M^i_{K_i})$ \texttt{//Pre-aggregation}
            \STATE Transfer knowledge from $M^i$ to corresponding proxy model $PM^i$ \texttt{//WIRE}
        \ENDFOR
        \STATE Aggregate proxy models $PM^G = FILE(PM^1, PM^2, \dots, PM^N)$ \texttt{//FILE}
        \FOR{each device type $i \in \{1,\dots,N\}$ in parallel}
            \STATE Transfer knowledge from global proxy model $PM^G$ to device model architecture $M^i$ \texttt{//WIRE}
        \ENDFOR
    \ENDFOR
\end{algorithmic}
\begin{flushleft}
\textbf{Device-side (for the $j$-th device of the $i$-th type):}
\end{flushleft}
\begin{algorithmic}[1]
    \FOR{$epoch = 1, 2, \dots, Ep$}
        \STATE Receive latest device model $M^i$ from server
        \STATE Train local model $M^i_j$ on local private dataset $d^i_j$
        \STATE Upload trained local model $M^i_j$ to the server
    \ENDFOR
\end{algorithmic}
\end{algorithm}

\subsection{Proxy Model Construction}
\label{sec:proxymodelcons}

In contrast to partial-model aggregation, full-weight aggregation requires finding weight-wise correspondences between heterogeneous models in order to achieve weight-to-weight aggregation.
The participation of thousands of devices makes it impractical to perform direct model transformations between every two devices due to the substantial computational overhead involved. To address this challenge, \tool adopts a novel design. After gathering all the heterogeneous device models $M$=$\{M^{i}_{j}\}$, where $i\in\{1,\dots,N\}, j\in\{1,\dots,K_N\}$, \tool constructs a series of homogeneous proxy models $PM=\{PM^1,PM^2,...,PM^N\}$ for each model architecture. 

In our design, the systematic approach to determine the architecture of the proxy model is to choose the same model capacity (i.e. the same architecture) as the largest device model. This design ensures that the proxy model can hold the knowledge of all device models. It is worth noting that, according to theories of knowledge transfer~\cite{gou2021knowledge,cho2019efficacy}, a more complex architecture for the proxy model does not impact the final performance of \tool. Instead, it only increases the computational overhead without any significant improvement in performance.

Continuing the process, the initial step involves performing a traditional aggregation (i.e., FedAvg~\cite{mcmahan2017communication}) for each type within the set of heterogeneous models, this can be expressed as $M^i = FedAvg(M^i_j)$, where $j \in \{1,\dots,K_i\}$. Then, by introducing the proxy model, \tool can transfer knowledge from each aggregated model $M^{i}$ to its proxy model $PM^i$ with the assistance of a shared public dataset $d_{public}$ on the server-side. Note that the shared public dataset setting is commonly seen in previous FL works, both in homogeneous FL\cite{mo2021ppfl} or heterogeneous FL\cite{itahara2020distillation,li2019fedmd,lin2020ensemble,cheng2021fedgems}. In \tool, the public dataset is stored and utilized on the server-side, thus avoiding any storage or computational overload on the devices.

Therefore, on a high level, with the PROM, the full-weight aggregation with heterogeneous models in \tool can achieve as follows. First, utilize the WIRE component to transfer knowledge from pre-aggregated models to their corresponding models as follows:
\begin{equation}
PM^i = WIRE(M^i), \text{where}\  i \in \{1,2,...,N\}
\end{equation}
Subsequently, FILE accomplishes the full-weight aggregation of heterogeneous models by aggregating the proxy models as:
\begin{equation}
PM^{G}=FILE(PM^1,PM^2,...,PM^N)
\end{equation}
Since the global model $PM^{G}$ remains heterogeneous with each device model, \tool utilizes WIRE once again to transfer the aggregated knowledge from the global model to $M^i$ for the purpose of updating:
\begin{equation}
M^i = WIRE(PM^{G}), \text{where}\ i \in \{1,2,...,N\}
\end{equation}
Lastly, we can deploy the $M^i$ to each device model $M^i_j, j\in\{1,\dots,K_N\}$ to achieve model update. Specifically, the WIRE and FILE are illustrated in the following sections.

\subsection{Weight-wise Knowledge Transfer}
\label{sec:wire}

\begin{figure}[htb]
    \centering
    \includegraphics[width=0.75\linewidth]{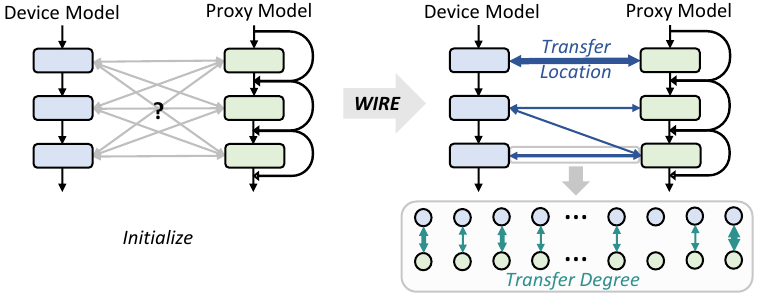}
	\caption{\revised{Design of WIRE. The width of the arrows represents the value of the transfer location/degree.}}
	\label{fig:wire}
\end{figure}

WIRE serves to establish a weight-wise knowledge transfer between heterogeneous models. The intuition is that full-weight aggregation is only feasible when the device models are transformed into homogeneous models, allowing for the establishment of head-to-head weight mapping. Therefore, it is essential and necessary to design a method that enables accurate and automatic model transformation. \revised{In our approach, we employ the meta-learning technique~\cite{sun2019meta,finn2017model} to learn transfer patterns at both the layer-to-layer (i.e., transfer location) and weight-to-weight (i.e., transfer degree) levels. The meta-learning technique enables us to learn a mapping function that effectively captures the relationships between the parameters of different model architectures. By doing so, the server can accurately identify how knowledge from one model can be transferred to another, ensuring that essential features and representations are preserved throughout the process. As illustrated in Figure~\ref{fig:wire}, by constructing meta-networks, we can precisely capture the transfer locations and degrees between each pair of layers, thereby facilitating accurate knowledge transfer among heterogeneous models.}

Specifically, take the transfer knowledge from $M^i$ to $PM^i$, we design two meta-networks for each model pair because the two networks focus on different transfer granularity:  $TL^i$ is to learn the layer-to-layer relationship, while $TD^i$ is to learn the weight-to-weight relationship. By learning $TL^i$ and $TD^i$, WIRE first matches the layers between the proxy model $PM^i$ and the pre-aggregated model $M^i$ with $TL^i$, then for matched layers, WIRE calculates the mappings between neurons with $TD^i$. Finally, WIRE transfers the knowledge from $M^i$ to $PM^i$ by minimizing the difference of each related transfer pattern.

Specifically, $M^i_m, PM^i_n$ denote the intermediate feature maps of the $m_{th}$ layer of $M^i$, $n_{th}$ layer of $PM^i$, respectively. We leverage $TL^i$ to \textit{learn} which layers of $M^i$ should be transferred to which layers of $PM^i$ automatically, which can be denoted as:
\begin{equation}
    \label{equation:tl}
    {Loc}^i_{m,n} = TL^{i,\phi}_{m,n}(M^i_m(d_{public}))
\end{equation} 
where $Loc$ denotes the importance of transferring knowledge from the $m_{th}$ layer of $M^i$ to $n_{th}$ layer of $PM^i$. $\phi$ denotes the parameters of the meta-network. The transfer location loss function can be defined as:
\begin{equation}
\begin{aligned}
    \label{equation:tl-loss}
     \mathcal{L}_{location}(\theta | d_{public}, \phi) = 
     \sum_{(m,n)} {Loc}^i_{m,n} \mathcal{L}_{degree}^{m,n}(\theta | d_{public}, Deg^i_{m, n})
\end{aligned}
\end{equation}
where $\mathcal{L}_{degree}$ refers to the transfer degree loss and $\theta$ consists of the parameters of the proxy model, and $Deg$ refers to the learnable transfer degree, which is explained below.

To obtain the precise transfer location, we require measuring the transfer degree in each location. Considering that not all parameters are equal in the device model, we take the feature map $M^i_m$ as the input of $TD^i$ and the softmax output as the channel weights to define the degree of knowledge transfer. Formally, the meta-network $TD^i$ is given as:
\begin{equation}
    \label{equation:td}
    {Deg}^i_{m,n} = TD^{i,\phi}_{m,n}(M^i_m(d_{public}))
\end{equation} 
The loss of transfer degree can be calculated as:
\begin{equation}
\begin{aligned}
    \label{equation:meta-loss-all}
    \mathcal{L}_{degree}^{m,n}(\theta | d_{public}, {Deg}^i_{m,n})  = 
      {Deg}^i_{m,n}(r_{\theta}(PM^{i,\theta}_{n}(d_{public})) - M^i_{m}(d_{public}))^{2}
\end{aligned}
\end{equation} 
where the $r_{\theta}$ represents a linear transformation parameters.

Thus, the final loss can be calculated as:
\begin{equation}
    \label{equation:final-loss}
    \mathcal{L}_{final} = \mathcal{L}_{CrossEntropy} + \mathcal{L}_{location}
\end{equation} 
By minimizing the $\mathcal{L}_{final}$, we can obtain the proxy model $PM^i$ with well-transferred knowledge. \revised{The cross-entropy loss $\mathcal{L}_{CrossEntropy}$ is also crucial in the $\mathcal{L}_{final}$, as it ensures that the proxy model inherits not only the final outputs but also the nuanced decision boundaries of the original model. This helps WIRE preserve critical knowledge during knowledge transfer, improving the overall accuracy and efficiency of MOSS.
The WIRE module is executed as Algorithm~\ref{algorithm:wire}.}

\begin{algorithm}[h]
\color{black}
\caption{\revised{The WIRE module}}
\label{algorithm:wire}
\begin{flushleft}
    \textbf{Input}: Pre-aggregated model $M^i$, proxy model $PM^i$ with parameter $\theta^i$, public dataset $d_{public}$
    
    \textbf{Output}: Proxy model $PM^i$ with accurately and automatically knowledge transferred
\end{flushleft}
\begin{algorithmic}[1] 

\STATE Initialize meta-networks $TL^i, TD^i$ with parameter $\phi$
\STATE Initialize $\mathcal{L}_{degree} = 0, \mathcal{L}_{location} = 0, \mathcal{L}_{CrossEntropy} = 0$
\FOR{\textit{epoch}=1, 2, \dots, \textit{Ep}}
\FOR{\textit{$(x,y) \in d_{public}$}}
\FOR{\textit{(m,n) $\in$ pairs}}
\STATE ${Loc}^i_{m,n} = TL^{i,\phi}_{m,n}(M^i_m(x))$
\STATE ${Deg}^i_{m,n} = TD^{i,\phi}_{m,n}(M^i_m(x))$
\STATE $\mathcal{L}_{degree}^{m,n} = {Deg}^i_{m,n}(r_{\theta^i}(PM^{i}_{n}(x)) - M^i_{m}(x))^{2}$
\ENDFOR
\STATE $\mathcal{L}_{location} = \mathcal{L}_{location} + \sum_{(m,n)} {Loc}^i_{m,n}(\mathcal{L}_{degree}^{m,n})$
\STATE $\mathcal{L}_{CrossEntropy} = \mathcal{L}_{CrossEntropy} + \mathcal{L}_{CrossEntropy}(\theta^i| x, y)$
\ENDFOR
\STATE $\mathcal{L}_{final} = \mathcal{L}_{CrossEntropy} + \mathcal{L}_{location}$
\STATE $\mathcal{L}_{final}$.backward()
\ENDFOR
\RETURN knowledge well-transferred proxy models $PM^i$ with updated parameter $\theta^i$.
\end{algorithmic}
\end{algorithm}

\subsection{Fidelity-guided Aggregation}
\label{sec:file}

An inefficient model aggregation process likewise slows down the convergence of FL. The most prominent issue is the neglect of transfer quality. For instance, when the learned correspondences between models are suboptimal, directly aggregating the knowledge-transferred proxy models might result in adverse effects, such as decreased accuracy, prolonged convergence times, and ultimately a significant rise in on-device overhead. To ensure the efficiency of aggregation, we introduce the concept of fidelity.

Fidelity, in this context, serves as a quantification of how well a proxy model aligns with the behavior of the pre-aggregated model. By considering fidelity, we ensure that the aggregated proxy models are of higher quality, possessing a more accurate and nuanced representation of the pre-aggregated model's performance. This strategic inclusion of fidelity into the aggregation process aims to enhance the overall effectiveness of the Federated Learning framework by mitigating the issues associated with subpar knowledge transfer and enhancing the final performance of the aggregated models. \revised{Intuitively, the fidelity score $Fid_{i}$ first measures the cosine similarity of the output logit between $M^i$ and $PM^i$ and then adjusted to a range of $[0,1]$, as:}
\revised{
\begin{equation}
\label{eq:fid1}
\begin{aligned}
    Fid_{i} = \frac{\cos(M^i(d_{public}),PM^i(d_{public})) + 1}{2}
\end{aligned}
\end{equation}}

Thus, we propose the fidelity-guided aggregation algorithm formally defined as Equation~\ref{eq:aggregation}, where $PM^{G}$ is the model possessing the knowledge of all proxy models, and $\alpha$ is a hyper-parameter to balance the two metrics. By employing this approach, FILE can assign higher weights to models that exhibit high fidelity, thereby leveraging their learned knowledge more effectively. Conversely, for models with low fidelity, we reduce their weights to mitigate their impact on the overall knowledge aggregation process. This enables FILE to prioritize the models that contribute positively to the overall accuracy and performance of the aggregated model.
\begin{equation}
    \label{eq:aggregation}
    \begin{aligned}
        PM^{G}=  FILE(PM^1,\dots,PM^N) =\sum_{i=1}^N \frac{Fid_i}{\sum_{i=1}^N Fid_i} \times PM^i
    \end{aligned}
\end{equation}

\section{Evaluation}\label{sec:eval}

This section evaluates the effectiveness of \tool to improve the model
performance and reduce computation cost under realistic settings. We choose the experimental settings that are close to the real-world scenario for heterogeneous FL. In this section, we aim to answer the following research questions:

\begin{tcolorbox}[size=small]

\noindent \textbf{RQ 1}: How is the convergence speed of \tool compared to other methods?

\noindent \textbf{RQ 2}: How much time does \tool save on devices?

\noindent \textbf{RQ 3}: How much energy can \tool save on devices?

\noindent \textbf{RQ 4}: How is the model performance of \tool compared with existing solutions?

\noindent \textbf{RQ 5}: How much transmission volume can \tool save on devices?
\end{tcolorbox}

In the following part of this section, we will first introduce the simulated
applications in Section~\ref{sec:applications}. Section~\ref{sec:setting} introduces the
experimental settings. Then, we perform a comprehensive evaluation of \tool
w.r.t other baselines and answer \textbf{RQ 1} in Section~\ref{sec:convergence}, \textbf{RQ 2} in Section~\ref{sec:time}, and \textbf{RQ 3} in Section~\ref{sec:energy}, \textbf{RQ 4} in Section~\ref{sec:acc}, and \textbf{RQ 5} in Section~\ref{sec:transmission}.

\subsection{Evaluated Applications}
\label{sec:applications}
We choose three representative real-world FL applications with
heterogeneous mobile devices. The selected applications are referred to the
production-level deployment of FL systems and widely used FL
benchmarks~\cite{pmlr-v162-lai22a,ouyang2021clusterfl,fedbalancer}. Our applications cover diverse input patterns and model
architectures. For each application, we adhere to the existing FL setup and simulate
the device users with non-IID data distribution~\cite{liu2021distfl,ouyang2021clusterfl}. We simulate the non-IID distribution because, compared with IID
distribution, it simulates the real-world setting and is more challenging for
FL. 

\revised{
For each application, we define $\mathcal{D}$ represents the dataset. We define $m$ types of devices and simulate a total of $n$ devices, so each type has $\frac{n}{m}$ devices (for simplicity we assume $n$ is divisible by $m$). For the $i$-th device, the local dataset is denoted as $d_i$, where $d_i \in \mathcal{D}$. 
Additionally, following the prior solutions~\cite{li2019fedmd, cheng2021fedgems}, we maintain a labeled public dataset $d_{public}$ that is used by the server. We
follow a strict constraint that the public dataset is not overlapped with all the devices' data, which satisfies $d_{public} \in \mathcal{D}$ and $d_{public} \cap d_i = \emptyset, \forall d_i \in \mathcal{D}$.
This is because during the real-world FL process, the devices' data
is highly sensitive and private, and the server cannot access the data. Thus the
server can only use a different public dataset. }

We introduce each application and how we simulate the data distribution
$\mathcal{D}$ in the following part. The detailed settings are also displayed in Table~\ref{tab:simulated_dataset}.

\begin{table*}[!htb]
\centering
\caption{The dataset settings of different applications.}
\setlength{\belowcaptionskip}{-4cm}
\label{tab:simulated_dataset}
\begin{adjustbox}{max width=1.0\linewidth}
\begin{tabular}{@{}cccccccccc@{}}
\toprule
Application & Dataset & Device index & Model & Distribution & \#Sample & Public distribution & \#Public sample \\ \midrule
 
\multirow{3}{*}{Image Classification} & \multirow{3}{*}{CIFAR-10~\cite{krizhevsky2009learning}} & 1-100 & Large & \multirow{3}{*}{Dirichlet} & \multirow{3}{*}{100} & \multirow{3}{*}{Dirichlet} & \multirow{3}{*}{100} \\
 &  & 101-200 & Medium &  &  &  &  \\
     &  & 201-300 & Small &  &  &  &  \\ \midrule
 
\multirow{3}{*}{Speech Recognition} & \multirow{3}{*}{Google Speech Command~\cite{speechcommandsv2}} & 1-100 & Large & \multirow{3}{*}{Dirichlet} & \multirow{3}{*}{200}  & \multirow{3}{*}{Dirichlet} & \multirow{3}{*}{200} \\
 &  & 101-200 & Medium &  &   &  &  \\
 &  & 201-300 & Small &  &   &  &  \\ \midrule
 
\multirow{3}{*}{Human Activities Recognition} & \multirow{3}{*}{Depth HAR~\cite{ouyang2021clusterfl,tu2021feddl}} & 1-10 & Large & IndoorDark & \multirow{3}{*}{50} & \multirow{3}{*}{Mixed} & \multirow{3}{*}{50} \\
 &  & 11-20 & Medium & IndoorNormal &  &  &  \\
 &  & 21-30 & Small & Outdoor &  &  &  \\ \bottomrule
 
\end{tabular}
\end{adjustbox}
\end{table*}

\subsubsection{Application 1: Image Classification}
\changes{
Image classification is one of the most critical applications for mobile devices equipped with cameras~\cite{pmlr-v162-lai22a,han2021legodnn}. Due to user preferences and heterogeneous deployment environments, data from various devices are often non-IID. Furthermore, these devices possess varying computational capacities due to differences in versions, brands, and hardware configurations, necessitating the deployment of heterogeneous models.}

Similar to other research on FL with heterogeneous models, we use the CIFAR-10 dataset~\cite{krizhevsky2009learning} to simulate this application. CIFAR-10 is a widely used dataset for image classification tasks in FL, containing 50,000 training images and 10,000 test images, with labels that span 10 classes. To emulate a typical scenario where each device has a highly heterogeneous data distribution, we simulated 300 devices, with devices 1-100 being high-end, 101-200 mid-range, and 201-300 low-end ($n=300$). We sampled 300 subsets each containing 100 training images ($m=300$) using a Dirichlet distribution with $\alpha=0.1$, commonly applied in FL data set partitioning, to simulate user routine data collection habits. \revised{Furthermore, to meet the needs of \tool, we sampled one more subset as the public dataset with 100 training images, while ensuring that there is no overlap with device data.} All images were centrally cropped to 32$\times$32 and normalized to zero mean for each channel.

\subsubsection{Application 2: Speech Recognition}
\changes{
Speech recognition is a typical application in smart homes and smartphones~\cite{yang2019proxitalk,sim2019investigation,pmlr-v162-lai22a}, integrated in numerous smart devices such as iPhone's Siri~\cite{applesiri}, and Amazon Alexa~\cite{amazonalexa}. In this task, the device takes voice as input and recognizes voice commands such as "start" or "stop."}

Similarly, we evaluated the performance of \tool in speech recognition using the Google Speech Command dataset~\cite{speechcommandsv2}, which is collected across various devices. This dataset includes 12 categories (10 specific words, "Unknown," and "None," totaling 12 labels), 85,511 training entries, and 4,890 test entries. \revised{We simulated the same number of devices as in the Image Classification application, totaling 300, with devices 1-100 as high-end, 101-200 as mid-range, and 201-300 as low-end ($m=3,n=300$). Likewise, to accommodate the highly heterogeneous device data distribution characteristic of modern FL research, we sampled 300 subsets each containing 200 training data using a Dirichlet distribution with $\alpha=0.1$. Following the previous method, we also sampleed an additional 200 training data as the public dataset.} For all voice data, we convert the voice to 32$\times$32 waveform images using MFCC~\cite{liu2015snooping} and normalize each channel to zero means.

\subsubsection{Application 3: Human Activity Recognition}
\changes{
Human activity recognition (HAR) is a typical task in home monitoring with mobile/IoT devices~\cite{ren2011robust,liu2020real} and is widely applied in FL research~\cite{tu2021feddl,ouyang2021clusterfl}. A challenging current dataset involves using depth images as input to recognize human activities.}

\changes{
We utilized the newer and widely used Depth HAR gesture recognition dataset in FL research on mobile/IoT devices~\cite{ouyang2021clusterfl,tu2021feddl}, which includes five different gestures (good, ok, victory, stop, fist) collected from three different environments (\textit{outdoor}, \textit{indoor dark}, and \textit{indoor}). The dataset was sampled in three different environments to provide non-IID data distributions with environmental differences. We simulated 30 devices (1-10 with the large model, 11-20 with the medium model, and 21-30 with the small model). \revised{We simulated the environment-difference non-IID setting, that randomly sampled 50 data from one environment of the dataset ($n=30$, $m=3$). For the public dataset, we randomly selected an additional 50 samples from the rest of the dataset across the three environments.} All images in the Depth HAR were resized to 32$\times$32.}

\subsection{Evaluation Setting}
\label{sec:setting}

\subsubsection{Baselines.} 
\changes{
We first conducted a comprehensive literature review on FL with heterogeneous model schemes. We explored key conferences and journals such as MobiCom, SenSys, TMC, SEC, NeurIPS, and others over the past years. We referenced a wide range of surveys~\cite{deng2022tailorfl,cho2022heterogeneous,liu2023recent} and endeavored to include all state-of-the-art (SOTA) solutions. Following the two categories of solutions we discussed—pruning-based and distillation-based solutions—we identified 12 papers representing SOTA solutions for FL with heterogeneous models. Although pruning-based solutions do not support FL with arbitrary heterogeneous model architectures, which deviates from our scenario on mobile/IoT devices, we still included seven baselines by pruning their original large model into smaller sizes that equal to other model architectures. We list them as follows:}

\begin{itemize}

    \item Hermes~\cite{li2021hermes}: Hermes presents a communication and inference-efficient FL framework designed for heterogeneity. It uses structured pruning to allow devices to learn personalized and structured sparse DNNs, reducing communication costs and improving efficiency.
    
    \item TailorFL~\cite{deng2022tailorfl}: TailorFL is a dual-personalized FL framework that customizes submodels for each device, accounting for both system and data heterogeneity. It employs a resource-aware pruning strategy and a scaling-based aggregation method, enhancing both the efficiency and accuracy of the FL process.
    
    \item FedMask~\cite{li2021fedmask}: FedMask introduces a communication and computation efficient framework of FL with heterogeneous models where devices learn a personalized and structured sparse DNN using a sparse binary mask. This approach minimizes communication bandwidth requirements and enhances the efficiency of model training and inference on mobile devices.
    
    \item LotteryFL~\cite{li2021lotteryfl}: LotteryFL leverages the Lottery Ticket Hypothesis to develop a personalized and communication-efficient framework of FL with heterogeneous models, reducing communication costs significantly by exchanging only key subnetworks between the server and clients.
   
    \item FedSkel~\cite{luo2021fedskel}: FedSkel optimizes FL with heterogeneous models on mobile/IoT devices by focusing updates on essential parts of the model, known as skeleton networks, thus achieving notable speedups and communication cost reductions with minimal accuracy loss.
    
    \item HeteroFL~\cite{diao2020heterofl}: HeteroFL addresses the challenge of training heterogeneous local models on devices with diverse computational and communication capabilities, enabling a unified global inference model without requiring uniform architecture across devices.
    
    \item FjORD~\cite{horvath2021fjord}: FjORD employs Ordered Dropout and self-distillation to handle system heterogeneity in FL with heterogeneous models, allowing for the extraction of tailored models according to device capabilities, significantly improving performance and maintaining model structure integrity.

\end{itemize}

For distillation-based solutions, which align more closely with our scenario, we included the following five baselines:

\begin{itemize}

    \item FedMD~\cite{li2019fedmd}: FedMD uses transfer learning and knowledge distillation within an FL framework with heterogeneous models to enable participants to contribute with independently designed models, enhancing collaborative learning.
    
    \item FedDF~\cite{itahara2020distillation}: FedDF explores ensemble distillation for model fusion in FL with heterogeneous models, allowing for flexible aggregation of heterogeneous device models and achieving faster training with fewer communication rounds compared to traditional FL methods.
   
    \item DS-FL~\cite{lin2020ensemble}: DS-FL is a distillation-based semi-supervised FL framework with heterogeneous models that reduces communication costs drastically by exchanging model outputs instead of parameters, enhancing model performance through data augmentation from unlabeled datasets.
    
    \item FedGEMS~\cite{cheng2021fedgems}: FedGEMS integrates knowledge transfer from multiple teacher devices to a server model in FL with heterogeneous models, enhancing both server and device model performance while maintaining robustness to security threats and efficiency in communication.
    
    \item Fed-ET~\cite{cho2022heterogeneous}: Fed-ET introduces an ensemble knowledge transfer method in FL with heterogeneous models, where small device models train a larger server model, effectively managing data heterogeneity and significantly outperforming traditional FL algorithms in terms of model generalization and communication efficiency.
    
\end{itemize}

For the baselines that provide publicly available code (\eg FedSkel, FedMD, HeteroFL, FjORD), we
directly use their implementation. For the approaches that do not provide public
code, we rigorously implement the techniques following the paper description.

\subsubsection{Models}
For each scenario and baseline, we simulate three levels of model heterogeneity
(\ie \texttt{Large}, \texttt{Medium}, and \texttt{Small}) to simulate a challenging yet realistic scenario~\cite{cho2022heterogeneous}.
During the FL process, all the three-level heterogeneous models engage in the
training procedure. We use the \texttt{Large} models to represent the complex models that
are run on high-end devices (\eg Huawei Mate X3~\cite{matex3}). The \texttt{Medium} models represent those run on the general-level devices (\eg Google Pixel 5~\cite{pixel5}). The \texttt{Small} models are assumed to run on low-end devices (\eg Huawei Harmony TV SE~\cite{huaweitv} whose memory size is only 1GB).

We select three representative heterogeneous models of different architectures
as the \texttt{Large}, \texttt{Medium}, and \texttt{Small} models for all the
scenarios. For the \texttt{Large} model, we select ResNet18~\cite{he2016deep} because it
is widely used in prior heterogeneous FL literatures~\cite{horvath2021fjord,cho2022heterogeneous,lai2021oort,pmlr-v162-lai22a}. For
\texttt{Medium} model, we select MobileNetV2~\cite{sandler2018mobilenetv2} because it is a
representative model that has been deployed to a large amount of mobile
devices~\cite{chen2023eefl}. For the \texttt{Small} model, we select LeNet~\cite{lecun1998gradient} because
it is one of the most lightweight models (the optimized number of parameters is only 5.7K). 
However, pruning-based solutions do not support aggregating models with different architectures because they must aggregate each layer. Thus, we use two smaller variants of ResNet18 as \texttt{Medium} and \texttt{Small}. Let $p \times$ResNet18 represent a ResNet18 model whose number of parameters is reduced to $p$ times of the original ResNet18. We use 0.4$\times$ResNet18 as the \texttt{Medium} model and 0.05$\times$ResNet18 as the \texttt{Small} model, to make the parameters of the two models essentially equal to those of MobileNetV2 and LeNet.

\subsubsection{Testbeds}
The FL server is a workstation equipped with two Nvidia RTX A6000 GPUs, Intel Xeon CPU with 24 cores, and 128GB memory. For the mobile devices, we use three devices with different compatibility to simulate the heterogeneous environment. Specifically, we use one Huawei Mate X3~\cite{matex3} running Harmony OS 4.0 as the high-level device with the large model, one Google Pixel 5~\cite{pixel5} running Android 12.0 as the general-level device with the medium model, and one Huawei Harmony Smart TV~\cite{huaweitv} running Harmony OS 2.0 as the low-end device with the small model. All the devices are equipped with AidLux~\cite{aidlux} to support Python and PyTorch libraries. A detailed description of the devices is shown in Table~\ref{tab:device}. In the evaluations, all devices are connected to an efficient WiFi network.

\begin{table}[!htb]
\centering
\caption{Devices for the testbed experiments.}
\begin{adjustbox}{max width=1\linewidth}
\begin{tabular}{cccccc}
\toprule
Device & Year & Platform & Clockspeed & Memory & Storage \\ \midrule
Huawei Mate X3 & 2023 & Snapdragon 8+ & 3.2GHz & 12GB & 512GB \\ \midrule
Google Pixel 5 & 2020 & Snapdragon 765G & 2.4GHz & 8GB & 128GB \\ \midrule
Huawei Harmony TV SE & 2022 & Arm Cortex A55 & 1.2GHZ & 1GB & 8GB \\
\bottomrule
\end{tabular}
\label{tab:device}
\end{adjustbox}
\end{table}

\subsubsection{Implementation Details}
\label{sec:implement}

\revised{
We implement \tool in Python 3.7 and PyTorch 1.8. Following the prior works on meta-learning~\cite{jang2019learning,pan2022distribution}, we construct the meta-networks of \tool as one-layer fully-connected networks for each layers-pair. For example, for the $m_{th}$ layer of $M^i$, $n_{th}$ layer of $PM_i$, the meta-network takes the feature of the $m_{th}$ layer as input and outputs the $Loc_{m,n}$ and $Deg_{m,n}$. As described in Section~\ref{sec:proxymodelcons}, we set the architecture of the proxy model as ResNet18, which is consistent with the \texttt{Large} model. 
For \tool and all the baselines, we use the same hyperparameters for a fair comparison. We use the SGD as the optimizer. The learning rate is set to 1e-3 with a momentum of 0.05. We set the batch size as 32, set the local training epoch to five, set the meta-learning-based knowledge transfer round in WIRE to five, and set the total federated rounds as 50 to guarantee convergence and report the final accuracy. All experiments are repeated by three times, and the average results are reported.}

\subsubsection{Metrics}
We report four metrics between \tool and all baselines: the number of FL rounds to convergence, the total time to complete
FL, the energy consumption of devices, and the accuracies for different devices. The number of FL
rounds is a quantitative metric to evaluate the convergence speed and is a
platform-independent metric. The total time to complete FL is a
platform-dependent metric that reports the authentic time cost on real devices.
The energy consumption evaluates the total energy cost of the FL process on real
devices. The accuracies represent the
final performance of the FL models for different devices.

\subsection{Convergence Rounds}
\label{sec:convergence}

\begin{figure*}[!tb]
    \centering
    \includegraphics[width=1\linewidth]{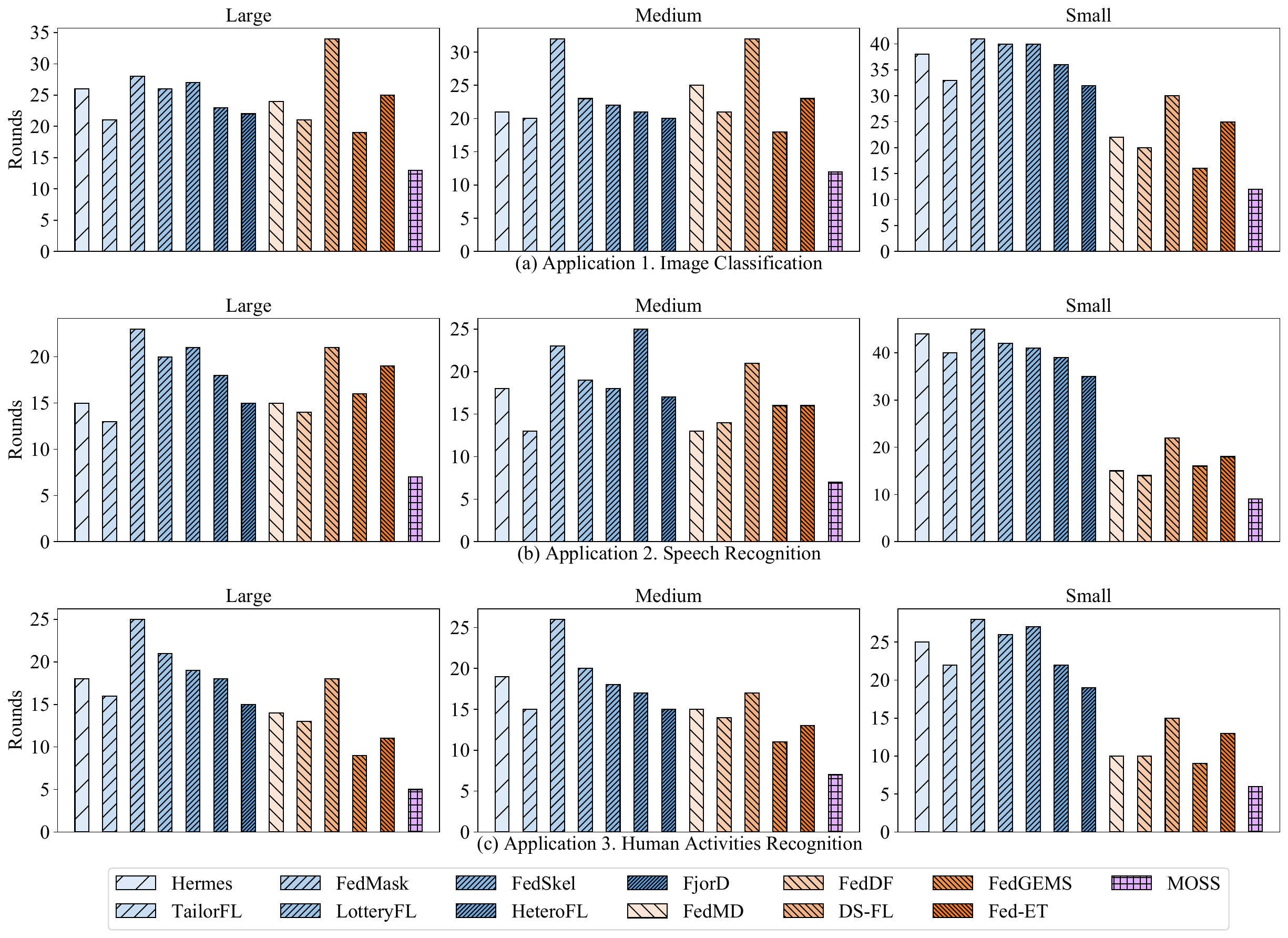}
	\caption{Comparison of FL rounds to achieve convergence.}
	\label{fig:flrounds}
 \vspace{-3ex}
\end{figure*}

In this section, we compare \tool and baselines w.r.t. the number of FL rounds to
achieve convergence. Recall that in Section~\ref{sec:motivation}, the motivation of \tool is
to improve the efficiency of transferring knowledge among heterogeneous models and
reduce the number of FL rounds. Thus we expect \tool can achieve faster
convergence speed. 
\revised{
For each case, we record the training process of all the device
models and take the number of convergence rounds as each device model reaches a stable accuracy where further training does not significantly improve the model's performance.~\cite{tabatabaie2024driver,li2024echopfl,liu2021distfl}}
The results are shown in Figure~\ref{fig:flrounds}. In the
figure, pruning-based methods are displayed in blue histograms with slash hatch, distillation-based are displayed
in orange histograms with back slash hatch, and \tool is represented by the
purple histogram with the cross. Each row of Figure~\ref{fig:flrounds} represents one
scenario and each column represents one type of model.

Generally speaking, we can observe that for both \tool and baselines,
\texttt{Small} models converge slower than \texttt{Large} models. For all cases,
the average convergence rounds of \texttt{Small} models is 26, which is 7 rounds
longer than those of \texttt{Large models} (19 rounds). It is because smaller models may require more rounds to fully utilize the aggregated knowledge~\cite{li2020train}.

Figure~\ref{fig:flrounds} shows that \tool reduces the amount of FL rounds by a
large margin. For the \texttt{Large}, \texttt{Medium}, and \texttt{Small}
models, the average converge rounds of \tool are 8, 9, and 9. Compared
with the baselines, \tool reduces the average convergence rounds by 39.9\%. Notably,
compared with FedMask, one of the pruning-based baselines, \tool speeds up the convergence by $3.5\times$
. 
Besides, we also notice that \tool can better accelerate the convergence speed
for \texttt{Small} models. Compared with \texttt{Large}, \tool reduces the
convergence round of \texttt{Small} models by $1.7\times$, which is $4.2\times$ faster than
other baselines. 
We deem the faster convergence, especially for \texttt{Small} models, comes from
the high quality of knowledge transfer at each round. The FL system does not
need to spend as many rounds as other baseline solutions to reach the
convergence state.

Other baselines' convergence speed is slow due to the ineffective
aggregation process. Averagely, the baselines have to spend 20, 19, and
25 for \texttt{Large}, \texttt{Medium}, and \texttt{Small} models to reach the
convergence, respectively. Although pruning-based
solutions provide a lightweight aggregation mechanism by preserving the mapping of the sub-model to the original model, they increase the convergence rounds of \texttt{Small} models by $4.1\times$ and $3.4\times$ than \tool, respectively. This is because the over-pruned sub-model will significantly lose important knowledge.

\subsection{Convergence Time on Real Devices}
\label{sec:time}

\begin{figure*}[!tb]
    \centering
    \includegraphics[width=1\linewidth]{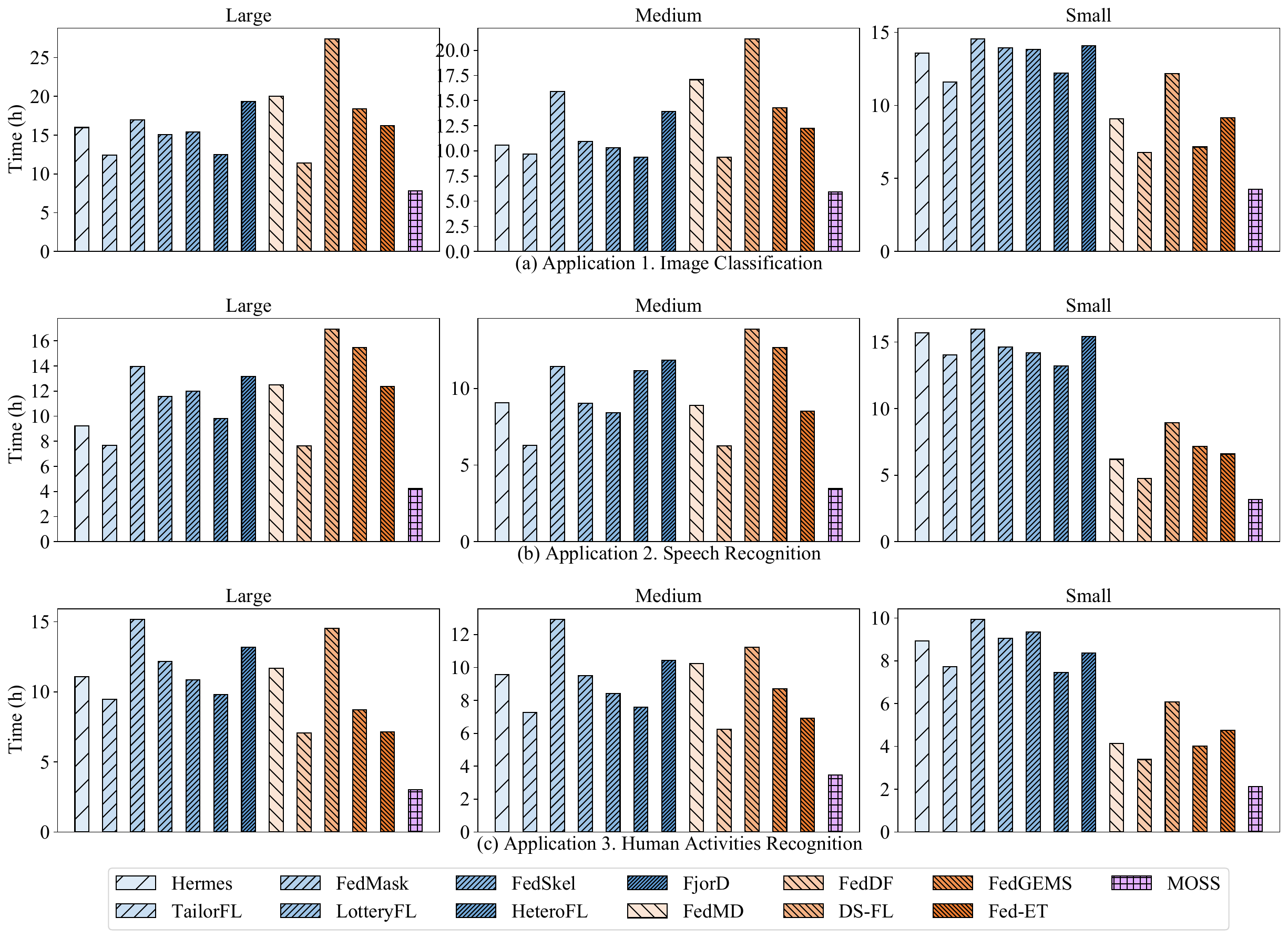}
	\caption{Comparison of the total time to complete FL for the devices.}
	\label{fig:traintime}

\end{figure*}

Besides the number of convergence rounds which is a platform-independent metric,
we also report the total convergence time on different devices to provide a
more authentic evaluation in real-world scenarios. To calculate the total
training time, we record the time cost per round on the devices and the
final time is computed as the convergence rounds multiplied by the time cost per
round. The time per round consists of three parts: local training, data
communication, and the waiting time for server response. We observe that the
local training time is the dominant part (over 90\%)~\cite{chen2023eefl}, but for a comprehensive evaluation we still report the total time.

The results are shown in Figure~\ref{fig:traintime} and the presentation style
is the same as Figure~\ref{fig:flrounds}. Compared with other baselines, \tool
has much smaller convergence time. For the \texttt{Large}, \texttt{Medium}, and
\texttt{Small} models, \tool takes 5.0 hours, 4.3 hours, and 3.2 hours to complete the training respectively.
Averagely, \tool reduces the convergence time by $62.9\%$. 

For the baselines, they have to spend 13.2 hours, 10.6 hours, and 9.9 hours to complete the training
procedure for the \texttt{Large}, \texttt{Medium}, and \texttt{Small} models.
DS-FL solution averagely spends 14.7 hours, which is the slowest compared with other
baselines. 
We observe that distillation-based solutions (orange bars) spend much more time
than other baselines on \texttt{Large} and \texttt{Medium} models (averagely by
$1.2\times$). It is because such solutions require an additional distillation process on
the devices, thus the local training time is much longer than other
baselines. In the additional evaluation, we notice that the local training time
of distillation solutions is around $1.4\times$ compared with other baselines. 
For the \texttt{Small} models, pruning-based solutions (blue bars) spend much
more time because their convergence time is much longer than other baselines (as
discussed in Section~\ref{sec:motivation}). It means such solutions are difficult to transfer knowledge to \texttt{Small} models.

\subsection{Energy Consumption}
\label{sec:energy}

\begin{figure*}[!tb]
    \centering
    \includegraphics[width=1\linewidth]{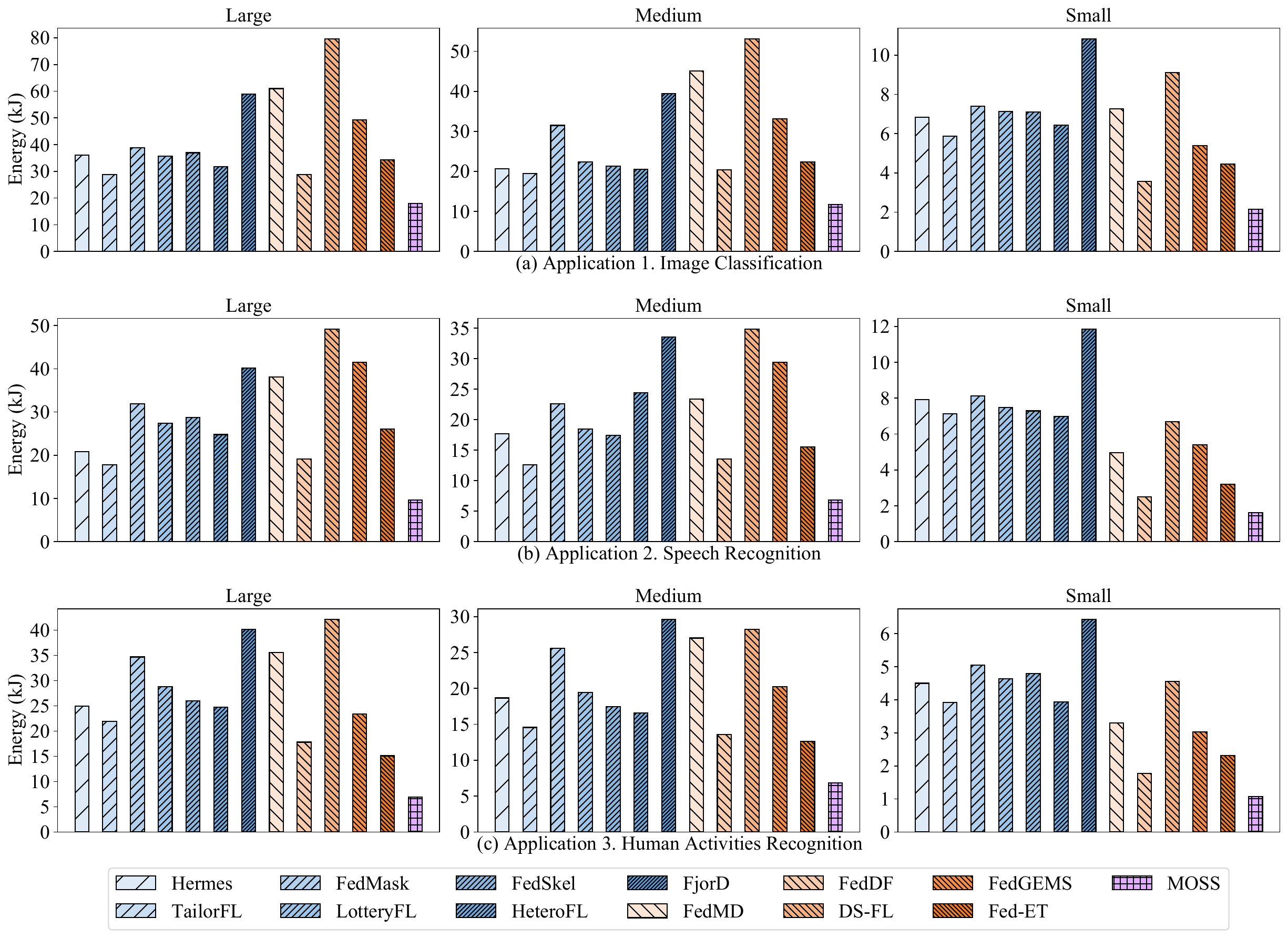}
	\caption{Comparison of the energy consumption to complete FL for the devices.}
	\label{fig:energy}
\end{figure*}

In this section, we report the energy consumption of \tool and baselines on the devices. For each case, we measure the consumed energy over five rounds,
compute the average energy consumption per round, and multiply the average
energy consumption by the convergence rounds to obtain the total energy
consumption during the FL process. The energy consumption is measured by the
system vFS (\texttt{/sys/class/power\_supply}) from the Battery Historian
tool~\cite{battery-historian}, which is widely used in the mobile community to measure the energy
consumption of mobile devices~\cite{xu2022mandheling}. This tool measures the energy consumption
of all the hardware components (\eg CPU, GPU, network module, etc.) and reports
the total energy consumption.

The results are shown in Figure~\ref{fig:energy}. \tool only consumes averagely
the least amount of energy (averagely 7.2 kJ). Compared with other baselines,
\tool consumes 6.1 times less energy. The standard battery of Huawei Mate X3
smartphones is a 5000mAh 3.7V battery, which can provide 66.6 kJ after full
charge. The energy consumption of \tool only accounts for 10.8\% of the battery.

Other baselines consume much more energy than \tool. Take distillation-based
solutions as an example. The maximum energy consumption on \texttt{Large} models
is 79.6 kJ, which is 1.2 times than the amount of energy that a Huawei
smartphone battery can provide. The large amount of consumed energy comes from
the increased convergence rounds (up to Figure~\ref{fig:flrounds} as displayed in Section~\ref{sec:convergence}).

\subsection{Accuracy}
\label{sec:acc}

\begin{table*}[]
\centering
\caption{Accuracy of three applications (\%). For each setting, we mark the highest accuracy in \textbf{bold}.}

\begin{adjustbox}{max width=1\linewidth}
\begin{tabular}{@{}ccccccccc|ccccc|cc@{}}
\toprule
&  & \multicolumn{7}{c|}{\ding{172} Pruning-based solutions} & \multicolumn{5}{c|}{\ding{173} Distillation-based solutions} &  \cellcolor[HTML]{E1AEFF}{} \\
 &  & \cellcolor[HTML]{deebf7}{Hermes} & \cellcolor[HTML]{c9def2}{TailorFL} & \cellcolor[HTML]{b3d1ec}{FedMask} & \cellcolor[HTML]{9ec4e7}{LotteryFL} & \cellcolor[HTML]{89b7e1}{FedSkel} & \cellcolor[HTML]{74abdc}{HeteroFL} & \cellcolor[HTML]{5f9dd7}{FjORD} & \cellcolor[HTML]{fbe5d6}{FedMD} & \cellcolor[HTML]{f8cbad}{FedDF} & \cellcolor[HTML]{f4b183}{DS-FL} & \cellcolor[HTML]{f0904e}{FedGEMS} & \cellcolor[HTML]{ed7c2f}{Fed-ET} & \multirow{-2}{*}{\cellcolor[HTML]{E1AEFF}{\tool}} \\ \midrule
\multirow{4}{*}{\rotatebox{90}{App. 1}} & \texttt{Large} & 70.5 & 74.1 & 69.7 & 68.8 & 65.3 & 71.2 & 77.8 & 70.1 & 67.2 & 66.1 & 68.5 & 67.3 & \textbf{80.1} \\
 & \texttt{Medium} & 69.4 & 71 & 68.4 & 67.5 & 63.6 & 69.8 & 71.2 & 64.6 & 61.5 & 58.4 & 67.8 & 66.9 & \textbf{74.5} \\
 & \texttt{Small} & 0.4 & 0.5 & 0.4 & 0.4 & 0.4 & 0.2 & 0.3 & 60.4 & 58.7 & 52.9 & 63.2 & 61.5 & \textbf{70.6} \\ \cmidrule(l){2-15} 
 & Average & 46.8 & 48.5 & 46.2 & 45.6 & 43.1 & 47.1 & 49.8 & 65.0 & 62.5 & 59.1 & 66.5 & 65.2 & \textbf{75.1} \\ \midrule
\multirow{4}{*}{\rotatebox{90}{App. 2}} & \texttt{Large} & 74.4 & 75.6 & 74.2 & 73.8 & 72.6 & 75.2 & \textbf{76.3} & 72.0 & 71.7 & 69.3 & 72.1 & 71.0 & \textbf{76.3}  \\
 & \texttt{Medium} & 71.4 & 72.7 & 71 & 70.7 & 70.4 & 72.2 & 73.3 & 69.3 & 69.9 & 68.6 & 70.2 & 70.4 & \textbf{73.5} \\
 & \texttt{Small} & 14.8 & 18.4 & 14.6 & 13.0 & 12.2 & 20.8 & 22.5 & 66.1 & 64.5 & 63.0 & 65.3 & 65.3 & \textbf{71.7} \\ \cmidrule(l){2-15} 
 & Average & 53.5 & 55.6 & 53.3 & 52.5 & 51.7 & 56.1 & 57.4 & 69.1 & 68.7 & 67.0 & 69.2 & 68.9 & \textbf{73.8} \\ \midrule
\multirow{4}{*}{\rotatebox{90}{App. 3}} & \texttt{Large} & 87.6 & 88.6 & 85.4 & 84.8 & 84.6 & 87.6 & 88.4 & 87.3 & 87.6 & 81.7 & 89.2 & 88.9 & \textbf{91.8} \\
 & \texttt{Medium} & 87.5 & 88 & 84.8 & 83.4 & 80.3 & 85.2 & 86.3 & 86.9 & 87.3 & 80.6 & 89 & 88.3 & \textbf{90.4} \\
 & \texttt{Small} & 33.5 & 34.6 & 28.3 & 27.5 & 28.8 & 23.5 & 24.3 & 83.8 & 82.2 & 76.8 & 86.8 & 86.5 & \textbf{89.1} \\ \cmidrule(l){2-15} 
 & Average & 69.5 & 70.4 & 66.2 & 65.2 & 64.6 & 65.4 & 66.3 & 86.0  & 85.7 & 79.7 & 88.3 & 87.9 & \textbf{90.4} \\ \bottomrule
\end{tabular}
\label{tab:acc}
\end{adjustbox}
\end{table*}

In this section, we report the accuracy of \tool and all the baselines. The
results are shown in Table~\ref{tab:acc}. For each scenario and baseline, we
report the average accuracy for \texttt{Large}, \texttt{Medium}, and
\texttt{Small} models, respectively. We also report the average accuracy of all
the models (the last row for each scenario). 

As we can observe from Table~\ref{tab:acc}, \tool consistently achieves the
highest accuracy among all applications when compared to existing solutions. For
the three applications, the average accuracies for \tool are 75.1\%, 73.8\%, and
90.4\%, respectively. \tool can outperform other baselines by 8.6\%, 4.6\%, and 2.1\%,
respectively. Moreover, the accuracy of each device does not deviate much from the average value. The standard deviation of the three applications is 1.2, 1.4, and 1.1 respectively. We do not display the per-device accuracy for brevity. The small deviation means that \tool can achieve stable performance for each device.

Besides, \tool can achieve high performance on \texttt{Small}
models. The average accuracy of \tool for all the \texttt{Small} models is 77.1\%,
which is only 5.6\% lower than the \texttt{Large} models and higher than other
baselines from 62.7\% to 5.4\%. It means the low-level devices (\eg TV) benefit most from \tool by
receiving a lightweight model with high accuracy. We believe the benefits to
\texttt{Small} models come from the full-weight aggregations. 
For \texttt{Large} models, \tool also achieves the highest accuracy. It is
because the full-weight aggregation of \tool can effectively transfer the
knowledge from \texttt{Small} models to \texttt{Large} models. Thus the
\texttt{Large} models can utilize the private data on low-level devices to
improve the accuracy.

For other baselines, they achieve decent results for all cases. For the three
applications, the average accuracy of all the baselines are 53.8\%, 60.2\%, and 74.6\%,
respectively. Even the highest accuracy among the baselines for each application
is lower than \tool by an average of 2.6\%. We also notice that transferring the
knowledge to \texttt{Small} models is difficult for these baselines, especially
for pruning-based solutions. In pruning-based solutions, the averaged accuracy of \texttt{Small} models is only 0.4\%, 16.6\%, and 28.7\%, respectively. The low accuracy is because of large amount of model reduction inevitably prunes out a significant amount of
crucial filters. The loss of crucial filters leads to degraded performance, and even model collapse~\cite{golkar2019continual,gordon2020compressing,fladmark2023exploring}.
In distillation-based solutions, take FedGEMS, which achieves the highest accuracy
among all the distillation-based solutions, as an example. The average accuracy
across the three applications for the \texttt{Small} model is only 71.7\%,
which is 5.4\% lower than \tool. The descent performance of the baselines
demonstrates the necessity to use full-weight aggregation to transfer knowledge of all layers among heterogeneous models.

\subsection{Transmission Volume}
\label{sec:transmission}

\begin{table*}[]
\caption{Comparison of cumulative transmission volume (MB).}
\setlength{\abovecaptionskip}{-0.4cm} 
\setlength{\belowcaptionskip}{-0.4cm}
\begin{adjustbox}{max width=1\linewidth}

\begin{tabular}{@{}ccccccccc|ccccc|cc@{}}
\toprule
&  & \multicolumn{7}{c|}{\ding{172} Pruning-based solutions} & \multicolumn{5}{c|}{\ding{173} Distillation-based solutions} &  \cellcolor[HTML]{E1AEFF}{} \\
 &  & \cellcolor[HTML]{deebf7}{Hermes} & \cellcolor[HTML]{c9def2}{TailorFL} & \cellcolor[HTML]{b3d1ec}{FedMask} & \cellcolor[HTML]{9ec4e7}{LotteryFL} & \cellcolor[HTML]{89b7e1}{FedSkel} & \cellcolor[HTML]{74abdc}{HeteroFL} & \cellcolor[HTML]{5f9dd7}{FjORD} & \cellcolor[HTML]{fbe5d6}{FedMD} & \cellcolor[HTML]{f8cbad}{FedDF} & \cellcolor[HTML]{f4b183}{DS-FL} & \cellcolor[HTML]{f0904e}{FedGEMS} & \cellcolor[HTML]{ed7c2f}{Fed-ET} & \multirow{-2}{*}{\cellcolor[HTML]{E1AEFF}{\tool}} \\ \midrule
\multirow{3}{*}{\rotatebox{90}{App. 1}} & \texttt{Large} & 1114.1 & 899.7 & 1199.7 & 1114.1 & 1156.8 & 985.4 & 942.6 & 15.8 & 899.7 & 21.3 & 11.9 & 1071.1 & 557.1 \\
 & \texttt{Medium} & 189.8 & 180.7 & 289.2 & 207.8 & 198.8 & 189.8 & 180.7 & 15.6 & 189.8 & 20.7 & 11.3 & 207.8 & 108.4 \\
 & \texttt{Small} & 11.6 & 10.1 & 12.5 & 12.2 & 12.2 & 11.3 & 9.8 & 13.8 & 6.1 & 18.8 & 10.2 & 7.6 & 3.7 \\ \midrule
\multirow{3}{*}{\rotatebox{90}{App. 2}} & \texttt{Large} & 640.8 & 555.4 & 982.6 & 854.4 & 897.1 & 769 & 640.8 & 0.7 & 598.1 & 1.3 & 0.8 & 811.7 & 299 \\
 & \texttt{Medium} & 157.4 & 113.7 & 201.1 & 166.1 & 157.4 & 218.6 & 148.6 & 0.6 & 122.4 & 1.3 & 0.8 & 139.9 & 61.2 \\
 & \texttt{Small} & 0.8 & 0.7 & 0.8 & 0.7 & 0.7 & 0.7 & 0.6 & 0.7 & 0.2 & 1.1 & 0.8 & 0.3 & 0.2 \\ \midrule
\multirow{3}{*}{\rotatebox{90}{App. 3}} & \texttt{Large} & 769.2 & 683.5 & 1068 & 897.1 & 811.7 & 769.2 & 640.8 & 0.7 & 555.4 & 0.9 & 0.4 & 469.9 & 213.6 \\
 & \texttt{Medium} & 166.1 & 131.1 & 227.3 & 174.9 & 157.4 & 148.6 & 131.1 & 0.7 & 122.4 & 0.8 & 0.5 & 113.7 & 61.2 \\
 & \texttt{Small} & 0.4 & 0.4 & 0.5 & 0.5 & 0.5 & 0.4 & 0.3 & 0.5 & 0.2 & 0.7 & 0.4 & 0.2 & 0.2 \\ \bottomrule
\end{tabular}
\label{tab:transmission}
\end{adjustbox}
\end{table*}

In this section, we evaluate the amount of data transmitted during the FL
process. Following prior work, we report the cumulative transmission volume
until convergence and the evaluation metric is MB. The results are shown in
Table~\ref{tab:transmission}.

For the three applications, the transmission volume of \tool is 223.1 MB, 120.1
MB, and 91.6 MB, respectively. In comparison, for the baselines, the average
transmission volume for the three applications is 318.1 MB, 227.5 MB, and 223.5
MB, respectively. It means \tool outperforms the baselines than 29.8\%, 47.2\%,
and 59.0\% for the three applications. It is worth noting that the advantage of
\tool is even more significant for \texttt{Small} models, which means low-end
devices benefits more from \tool, as discussed Section~\ref{sec:acc}.
Someone may observe that the distillation-based solutions (FedMD, DS-FL, and
FedGEMS) have low transmission volume. It is because they only transmit the
output logits on the shared public dataset rather than the device model.
However, such solution significantly increases the on-device computation
overhead and result in high energy consumption, as shown in ection~\ref{sec:energy} and Figure~\ref{fig:energy}.

\subsection{Ablation Study}
\label{sec:ablation}

\revised{
In this section, we conducted several experiments of ablation study to evaluate the necessity of three key modules, PROM, WIRE, and FILE.}

\subsubsection{Necessity of PROM}
\label{sec:ablation_prom}

\begin{figure}[h]
    \centering
    \includegraphics[width=1\linewidth]{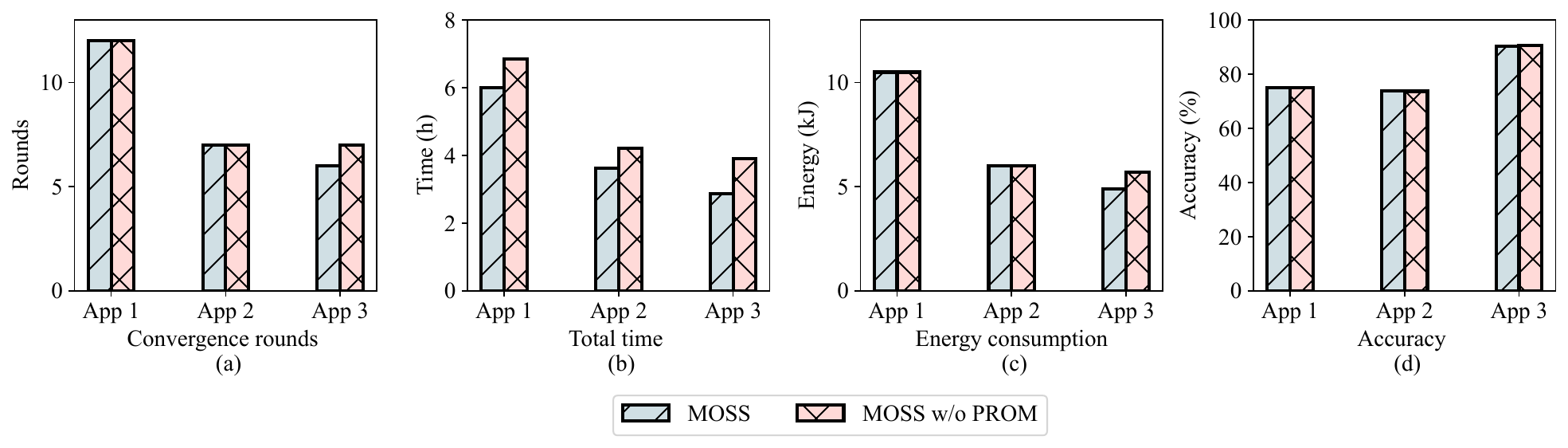}
	\caption{\revised{Comparison on Performance between \tool and \tool without PROM.}}
	\label{fig:albationPROM}
\end{figure}

\revised{
The construction of the proxy model in \tool is essential, as it reduces the necessary $n^2$ times knowledge transfers on the server to $2n$ times without sacrificing accuracy, thereby enhancing overall efficiency. To demonstrate the necessity of PROM, we conducted experiments without constructing proxy models, referred to as \tool w/o PROM. In this setup, for updating the $i$-th type of device model, we directly transformed the other pre-aggregated models to match its architecture using WIRE, requiring a total of $n^2$ times knowledge transfers.}

\revised{
In Figure~\ref{fig:albationPROM}, we present a performance comparison between \tool and \tool w/o PROM. The results show that while the convergence rounds and accuracies are roughly equivalent, \tool w/o PROM requires an additional round in application 3 compared to \tool. Furthermore, the total time increases by 14.2\%, 16.6\%, and 36.2\% across the applications, while energy consumption rises by 16.3\% in application 3. These findings clearly demonstrate that using PROM not only effectively reduces total time and energy consumption but also maintains convergence efficiency and overall accuracy.}

\subsubsection{Necessity of WIRE}
\label{sec:ablation_wire}

\begin{figure}[h]
    \centering
    \includegraphics[width=1\linewidth]{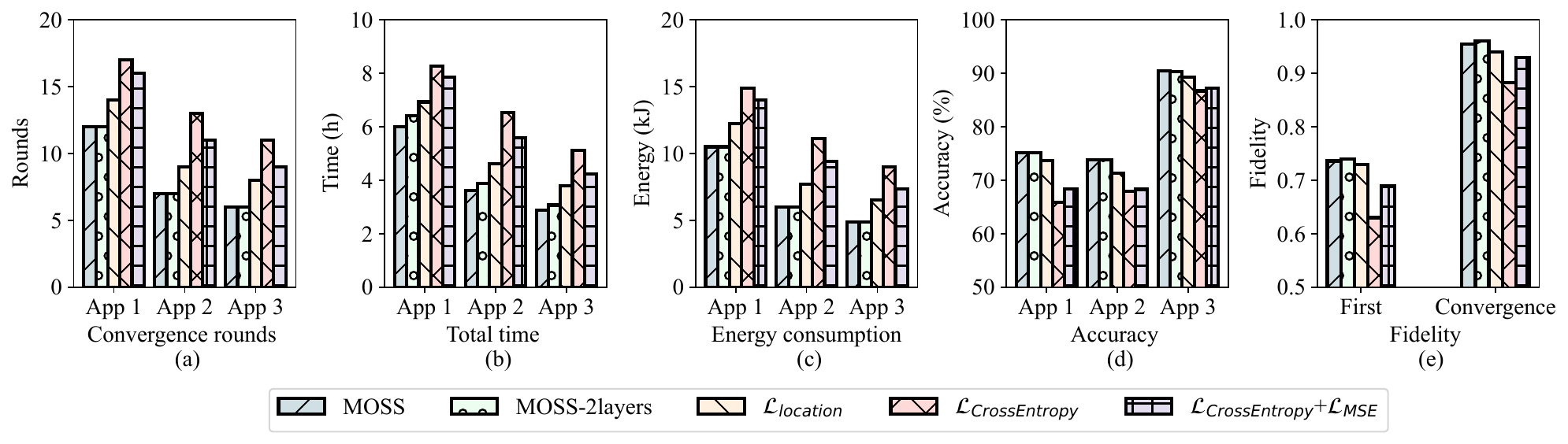}
	\caption{Necessity of WIRE.}
	\label{fig:albationWIRE}
\end{figure}

WIRE is a critical component of the Moss framework, facilitating efficient knowledge transfer between heterogeneous models. To assess the necessity and contributions of WIRE, we conducted a series of experiments comparing the performance of \tool with several variations: \tool-2layers (which employs more complex meta-networks), $\mathcal{L}_{location}$ (which uses only the meta-learning loss without the cross-entropy loss), $\mathcal{L}_{CrossEntropy}$ (which employs only the cross-entropy loss without meta-learning), and $\mathcal{L}_{CrossEntropy}+\mathcal{L}_{MSE}$ (which combines cross-entropy and mean squared error (MSE), replacing meta-learning with MSE). The evaluation metrics included the averaged convergence rounds, total time, energy consumption, accuracy, and fidelity. Fidelity is crucial for assessing the effectiveness of knowledge transfer. We compare the fidelity on the first FL round and the convergence round. By examining these variations, we aimed to highlight the efficiency and effectiveness of the WIRE module, as well as the roles played by the meta-learning and cross-entropy loss in the knowledge transfer process.

Figure~\ref{fig:albationWIRE} presents the complete experimental results. From these results, we observe that compared to $\mathcal{L}_{location}$, $\mathcal{L}_{CrossEntropy}$, and $\mathcal{L}_{CrossEntropy}+\mathcal{L}_{MSE}$, \tool demonstrates significant improvements. Specifically, on average, \tool can reduce convergence rounds by up to 40.3\%, total time by 38.7\%, and energy consumption by 40.4\%, while achieving an increase of 8.9\% in accuracy. These findings underscore the significance and effectiveness of meta-learning, while also highlighting the indispensable role of cross-entropy in the knowledge transfer process.

Additionally, when compared to \tool-2layers, \tool maintains similar performance in convergence rounds, energy consumption, accuracy, and fidelity,  with even slight improvements. This may be attributed to the increased complexity of the meta-networks, which can be more challenging to train. Furthermore, since larger meta-networks require longer computation times on the server, \tool is able to maintain a faster total time. Therefore, choosing a one-layer fully-connected network in \tool strikes an effective balance between accuracy and efficiency.

\subsubsection{Necessity of FILE}
\label{sec:ablation_file}

\begin{figure}[h]
    \centering
    \includegraphics[width=1\linewidth]{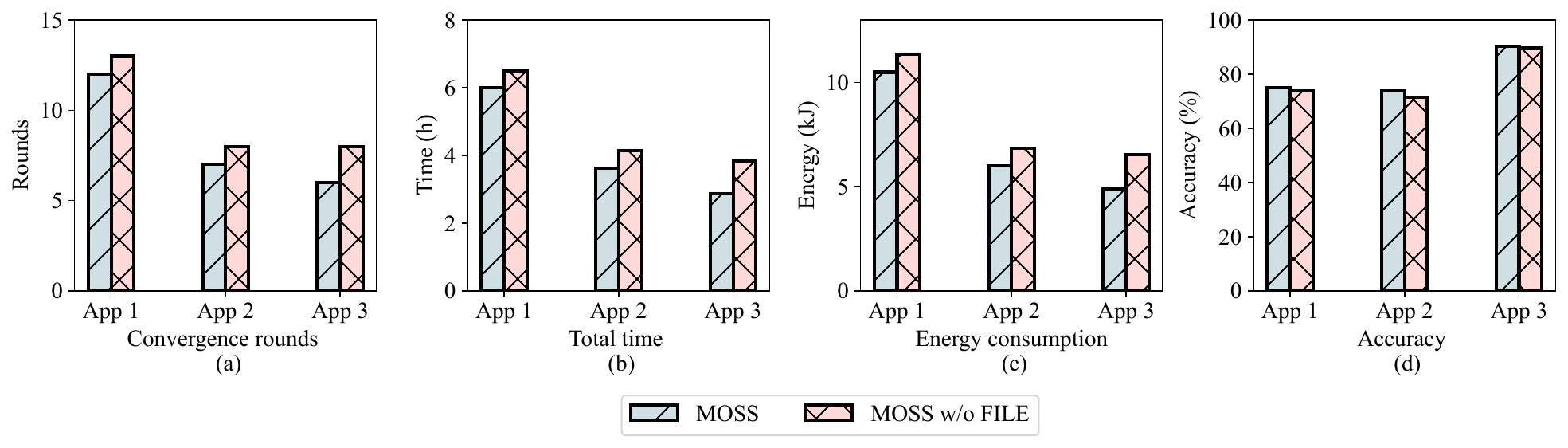}
	\caption{Comparison of the four metrics between aggregation with FILE and without FILE.}
	\label{fig:albationFILE}
\end{figure}
The FILE module is crucial for enhancing the performance of the \tool framework by ensuring that model updates are aggregated based on their fidelity. To evaluate the necessity of FILE, we compared the performance of \tool with that of a baseline approach using conventional FedAvg for model aggregation (denoted as \tool w/o FILE).

As shown in Figure~\ref{fig:albationFILE}, we can observe that \tool, utilizing FILE, achieved significantly better results across multiple metrics compared to the baseline. Specifically, \tool demonstrated improvements in convergence rounds, total time, energy consumption, and accuracy, with reductions of 15.1\% in rounds, 15.0\% in total time, and 15.3\% in energy consumption, while increasing accuracy by 2.0\%. These results highlight the effectiveness of FILE in prioritizing model updates that contribute the most to overall performance, thereby improving the efficiency of the aggregation process. In contrast, the baseline method, which treats all model updates equally, often leads to suboptimal performance.

\subsection{Robustness}
\label{sec:robustness}

In this section, we conducted several experiments under different settings to evaluate the robustness of \tool: performance under unrelated public dataset, performance under differential privacy settings, the impact of different fractions of participating devices and the impact of the public dataset size.

\begin{figure}[h]
    \centering
    \includegraphics[width=1\linewidth]{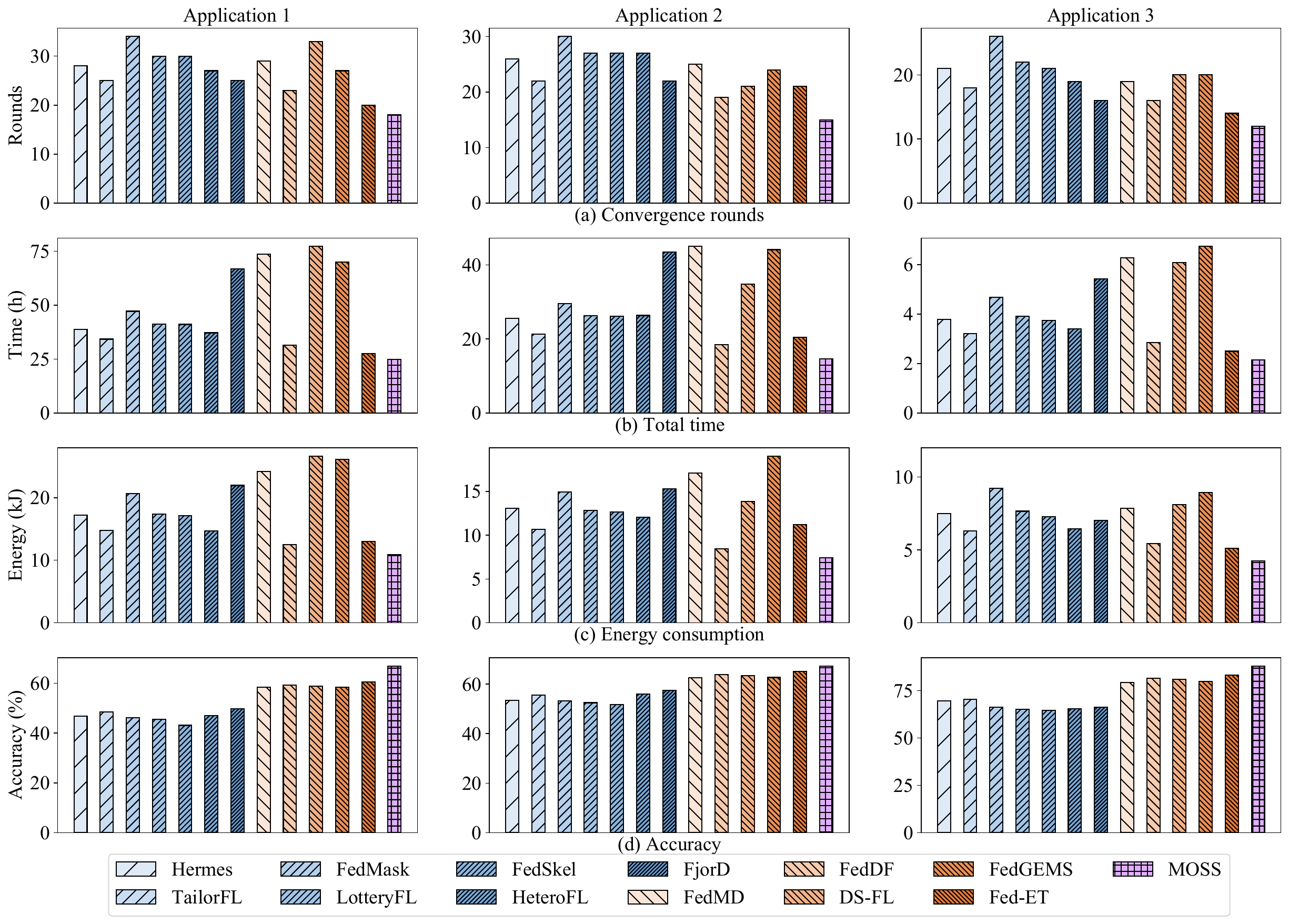}
	\caption{\revised{Evaluation on baselines and \tool with the unrelated public dataset in three applications.}}
	\label{fig:differntpublic}
\end{figure}

\subsubsection{Robustness Under Unrelated Public Dataset.}
\label{sec:robust_unrelated}
In existing works on FL with heterogeneous models that achieve model aggregation utilizing public datasets, methods such as FedMD~\cite{li2019fedmd} and FedGEMS~\cite{cheng2021fedgems} employ labeled public datasets from the same task but do not overlap with the device data. While DS-FL~\cite{itahara2020distillation} and Fed-ET~\cite{cho2022heterogeneous} do not require labels from the public dataset, they still expect the public dataset to be from the same task domain as the device data. Only FedDF~\cite{lin2020ensemble} indicates that it can handle scenarios where the public dataset is collected from unrelated tasks. This is because they assumed a more extreme case, where the server cannot obtain a public dataset related to the task at all. To demonstrate the robustness of \tool, we conducted additional experiments using public datasets unrelated to the device data.

Specifically, for the image classification task (Application 1), we used CIFAR-100~\cite{krizhevsky2009learning} as the public dataset, randomly sampling 100 images, distinct from the original CIFAR-10 training data. In the speech recognition task (Application 2), we employed the spoken digits dataset~\cite{jackson2018jakobovski} and sampled 200 data points, ensuring no overlap with the Google Speech Commands dataset. For the human activity recognition task (Application 3), we used the MSRA Hand Gesture dataset~\cite{qian2014realtime}, and randomly sampled 50 data, separate from the Depth HAR dataset. These unrelated public datasets were used exclusively on the server side, with the same sample size as the previously used public datasets, ensuring consistency in experimental setup while validating Moss's adaptability across varied data environments.

The experimental results show that Moss maintains its strong performance even when the public dataset is entirely unrelated to the device data, further highlighting its robustness in FL with heterogeneous models settings. Specifically, as shown in Figure~\ref{fig:differntpublic}, experimental results demonstrate that \tool can effectively reduce convergence rounds by 10.0\% to 47.1\% in the image classification task, 21.1\% to 50.0\% in the speech recognition task, and 14.3\% to 53.8\% in the human activities recognition task. In three applications, the total training time is improved by an average factor of 1.2$\times$ to 3.1$\times$, 1.3$\times$ to 3.0$\times$, and 1.2$\times$ to 3.1$\times$, and energy consumption is reduced by 1.2$\times$ to 2.5$\times$, 1.3$\times$ to 3.0$\times$, and 1.2$\times$ to 3.1$\times$. Additionally, \tool achieves accuracy improvements of 10.6\% to 55.2\%, 3.2\% to 30.2\%, and 5.5\% to 36.1\% for the respective three applications. Despite the difference in task domains between the public and device datasets, \tool continues to achieve state-of-the-art performance, reinforcing its capability to generalize effectively under unrelated public dataset settings.

\begin{figure}[h]

    \centering
    \includegraphics[width=1\linewidth]{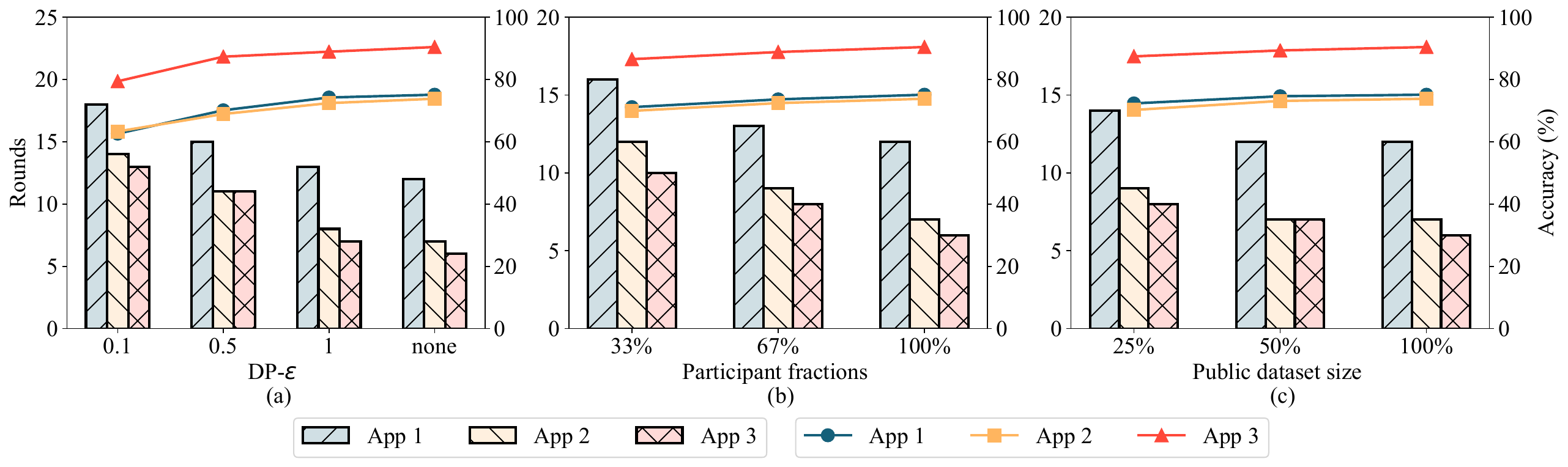}
	\caption{\revised{(a) Comparison of different DP settings. (b) Comparison of different participant fractions. (c) Comparison of the size of public dataset.}}
	\label{fig:robustness}
\end{figure}

\subsubsection{Robustness Under Differential Privacy (DP)}
\label{sec:robust_dp}

\revised{
Differential privacy (DP) has been widely researched in the context of FL as a defense mechanism to enhance the protection of device models and data privacy~\cite{yu2020salvaging}. While our study operates orthogonally to DP, we acknowledge that the noise introduced by DP can impact overall performance. Therefore, we aim to evaluate the robustness of \tool under differential privacy conditions. Following the prior research~\cite{liu2021distfl}, we set the DP hyperparameter $\epsilon$ to $0.1$, $0.5$, $1$, and \textit{none}, to evaluate the overall accuracy and the convergence round of \tool in three applications.}

\revised{
We demonstrate the results in Figure~\ref{fig:robustness} (a). We used the bars to represent the convergence rounds in three applications, while used the curves to represent the accuracy.
As $\epsilon$ decreases, indicating a higher level of privacy protection, we observed a corresponding decline in both convergence rounds and accuracy. Specifically, when $\epsilon=0.1$, the convergence rounds increased by 79.9\%, while accuracy decreased by 14.2\%. Nevertheless, these results demonstrate that \tool can still deliver competitive performance even under stringent privacy conditions. These findings highlight the robustness of \tool, showcasing its ability to maintain effective performance while adhering to higher privacy standards, making it suitable for deployment in privacy-sensitive environments.}

\subsubsection{Robustness Under Different Fractions of Participating Devices.}

In real-world deployment, not all devices may participate in the whole FL
process due to the network or battery issues. At each communication round, it is
highly possible that only a fraction of devices can participate in the FL
process. Therefore we evaluate how this factor will affect the performance of
\tool. In our evaluation, we compare three configurations of the participating
fractions in each communication round: 33\%, 67\%, and 100\%. For each
configuration, at each round, we randomly select the corresponding fraction of
devices to train the local models and transmit the model to the server.

The results are shown in Figure \ref{fig:robustness} (b), where the curves
represent the final accuracy of \tool and the bars represent the convergence
round. We can observe that as the fractions of participating devices increase,
the accuracy is improved. Specifically, the average accuracy for the three
configurations is 75.8\%,78.3\%, and 79.8\%, respectively. Meanwhile, the
convergence round decreases as the participant fraction increases. When the
fraction is 33\%, the convergence round is 13. As the fraction increased to 100\%,
the convergence round is 8. It means that \tool is robust to the change of
participant fraction and can achieve high performance even when only 33\% of
devices participate in each round.

\subsubsection{Robustness Under Different Public Dataset Size.}

The size of the public dataset on the server side is another important factor that
may affect the robustness of \tool. Someone may doubt that the high performance
of \tool comes from the large public dataset, or the requirement of the public
dataset may impede the scalability of \tool.
To study the impact of this factor, we select three configurations on the size
of the public dataset: 25\%, 50\%, and 100\%. The results are shown in Figure
\ref{fig:robustness} (c). We can observe that the average accuracy for the three
configurations are 76.6\%, 78.9\%, and 79.8\%, respectively. The convergence
rounds are 10, 9, and 8. It means that, although the size of the public dataset
may affect the performance of \tool, the impact is not significant. \tool does
not rely on large public dataset to achieve high performance. Besides, the
requirement of public dataset does not impede the deployment of \tool because
even if the server can only collect 25\% of the data (up to 50 samples for the
three applications), \tool can still achieve high performance.

\section{Related Work}
\label{sec:relatedwork}
The closest work to this paper is FL with heterogeneous models, which trains models of
different architectures to fit the diverse computation capabilities on mobile
devices. We
divided the related works in FL with heterogeneous models into three categories: pruning-based solutions, distillation-based solutions, and alternative solutions.

\textbf{Pruning-based
solutions.} LotteryFL~\cite{li2021lotteryfl}, Hermes~\cite{li2021hermes}, FedMask~\cite{li2021fedmask}, TailorFL~\cite{deng2022tailorfl}, FedSkel~\cite{luo2021fedskel}, HeteroFL~\cite{diao2020heterofl}, and FjORD~\cite{horvath2021fjord} preferentially consider the constrained computation ability on the device and reducing the devices' model size to fit the computation ability. Specifically, these solutions first
initialized an integral model architecture on the server side and customized a
smaller model for each device based on the device's computational capability.
The small model contains a subset of the filters from the original model. During
the training phase, the device only trains the small model, and the server only
aggregates the small model. However, they face critical limitations. First, these solutions require a consistent original model for effective pruning, which complicates the aggregation process across heterogeneous devices that may have customized architectures. Second, excessive pruning can lead to significant performance degradation if too many essential parameters are removed, disrupting model architecture and preventing convergence. This makes pruning-based methods unsuitable for real-world applications on mobile and IoT devices.

\textbf{Distillation-based
solutions.} FedMD~\cite{li2019fedmd}, FedDF~\cite{lin2020ensemble}, DS-FL~\cite{itahara2020distillation}, FedGEMS~\cite{cheng2021fedgems}, and FED-ET~\cite{cho2022heterogeneous} uses homogeneous logits as the proxy to transfer knowledge.
Such solutions allow devices to deploy arbitrarily defined model architectures.
During the training phase, the device first trains the local model and uses a
shared public dataset to calculate the predicted logits of the local model.
These logits are uploaded to the server and aggregated with other devices'
logits. The aggregated logits are distributed to the devices and used to
supervise the training procedure of the subsequent round. While these methods provide flexibility in model architectures, they encounter several limitations. Notably, the aggregation process often results in the loss of valuable information from intermediate layers, which are crucial for effective model performance. Research indicates that knowledge embedded in intermediate layers contains rich feature representations that contribute to a model's understanding of complex data~\cite{romero2014fitnets,heo2019comprehensive,tung2019similarity}. However, the reliance on global pooling layers during aggregation reduces the dimensionality of outputs, capturing only limited class knowledge and discarding detailed information~\cite{9340578,cho2019efficacy}. This loss of intermediate knowledge prolongs training rounds and can significantly impair model accuracy, particularly in resource-constrained environments where maintaining high performance is critical. Thus, while distillation-based solutions are adaptable, their effectiveness can be compromised by these inherent limitations.

\textbf{Alternative
solutions.} HAFL-GHN~\cite{litany2022federated} employs a Graph HyperNetwork to integrate diverse device architectures while maintaining privacy. However, it struggles to establish correspondences between different architectures, opting not to aggregate unmatched layers, which complicates real-world applications. DISTREAL~\cite{rapp2022distreal} introduces a distributed resource-aware learning strategy that allows clients to dynamically omit certain filters. While offering flexibility, it requires an initial comprehensive model, limiting its use in highly heterogeneous environments. FEDHM~\cite{fan2023model} decomposes model parameters into shared and individual components to reduce communication overhead. However, it incurs significant computational demands and relies on shared parameters that exceed the capabilities of low-end devices, restricting its adaptability in diverse settings. Overall, these alternative solutions also face challenges that limit their practical implementation in mobile/IoT environments.

Besides FL with heterogeneous models, researchers also study FL with heterogeneous data, which aims to improve the accuracy of the final model when the devices' data is different~\cite{sim2019investigation,mansour2020three,liang2020think,wang2020advances,reddi2021adaptive,cai2024famos}. Despite their effectiveness, these approaches fundamentally differ from \tool since they address a different problem. Such solutions are orthogonal to \tool and can be combined with \tool to further improve the performance of \tool on highly diverse data distribution.

\section{Discussion}
\label{sec:discussion}

We need to highlight that \tool maintains the standard practice of FL by only uploading the device model to the server, without transferring any raw data and extra information from devices. This is in compliance with data privacy regulations and ensures that no additional risk is introduced compared to conventional FL frameworks.

Meanwhile, while our method does not inherently increase data transmission, it is important to consider potential privacy leakage from model updates. To address this, for example, differential privacy (DP)~\cite{wei2020federated, liu2021distfl,hu2023federated} can be employed as a defense mechanism. DP works by adding noise to the model updates, thereby protecting individual data points from being inferred through the aggregated results. In our experiments, we follow the prior work to include an evaluation of Moss under DP conditions. Despite some impact on model accuracy due to the added noise, Moss continues to achieve SOTA performance, demonstrating its robustness even under privacy-preserving constraints. In addition, multi-party computation (MPC) techniques~\cite{gehlhar2023safefl,stevens2022efficient,chang2023privacy} are often used in FL to protect model aggregation against untrusted parties. \tool is compatible with MPC because Moss performs aggregation at the weight level without changing the aggregation rule in standard FL.

In summary, our approach respects the fundamental principles of FL by adhering to standard transmission and aggregation practices and is compatible with privacy-preserving enhancements like DP and MPC. This ensures that Moss remains a secure and effective solution for FL with heterogeneous models.

\section{Conclusion}
In this paper, we propose \textit{full-weight aggregation} with heterogeneous models to improve the accuracy and training efficiency of FL models. Our approach transfers heterogeneous models to homogeneous models and achieves full-weight aggregation. We design and implement a framework called \tool that aggregates heterogeneous models at a weight level. Our experiments on designed applications show that \tool outperforms other state-of-the-art methods by increasing accuracy by up to 8.6 percentage points, decreasing training time by $62.9\%$, and reducing energy consumption by up to $6.1\times$.

\section*{Acknowledgments}

We want to thank the anonymous reviewers for their valuable feedback. This work was partly supported by the National Science and Technology Major Project of China (2022ZD0119103) and the Beijing Natural Science Foundation (L243010). 

\bibliographystyle{ACM-Reference-Format}

\bibliography{references}

\end{document}